\documentclass[lettersize,journal]{IEEEtran}
\usepackage{amsmath,amsfonts}
\usepackage{array}
\usepackage[caption=false,font=normalsize,labelfont=sf,textfont=sf]{subfig}
\usepackage{textcomp}
\usepackage{stfloats}
\usepackage{url}
\usepackage{verbatim}
\usepackage{xcolor}
\usepackage{graphicx}
\usepackage{cite}
\usepackage{fontawesome5}
\usepackage{booktabs} 
\usepackage{multirow} 
\usepackage{graphicx} 
\usepackage[linesnumbered,ruled,vlined]{algorithm2e}
\usepackage{cite}
\usepackage[table]{xcolor}
\definecolor{best}{RGB}{245, 220, 190}
\definecolor{second}{RGB}{230, 225, 220} 
\makeatletter
\g@addto@macro{\UrlBreaks}{\do\-\do\/\do\_}  

\begin{document}

\title{Real-World Scene Recovery for Scattering-Degraded Images Using Spatial and Frequency Priors}


\author{Yun Liu, Tao Li, Guanghui Yue, Wenqi Ren, Cosmin Ancuti, Weisi Lin,~\IEEEmembership{Fellow, IEEE}
\thanks{This work was supported in part by the National Natural Science Foundation of China under Grant 62301453.  }
\thanks{Yun Liu and Tao Li are with College of Artificial Intelligence, Southwest University, Chongqing 400715, China (e-mail: yunliu@swu.edu.cn and lt3088919588@email.swu.edu.cn) }
\thanks{Guanghui Yue is with the School of Biomedical Engineering, Shenzhen University Medical School, Shenzhen University, Shenzhen 518060, China (yueguanghui@szu.edu.cn).}
\thanks{Wenqi Ren is with School of Cyber Science and Technology, Shenzhen Campus, Sun Yat-sen University, Shenzhen 518107, China (renwq3@mail.sysu.edu.cn) }
\thanks{Cosmin Ancuti is with the ETcTI, University Politehnica Timisoara, Timisoara 300006, Romania (cosmin.ancuti@upt.ro).}

\thanks{Weisi Lin is with College of Computing and Data Science, Nanyang Technological University (NTU), Singapore 639798, Singapore (wslin@ntu.edu.sg) }
}

\markboth{Journal of \LaTeX\ Class Files,~Vol.~14, No.~8, August~2021}%
{Shell \MakeLowercase{\textit{et al.}}: A Sample Article Using IEEEtran.cls for IEEE Journals}


\maketitle

\begin{abstract}
Scene recovery from real-world images degraded by scattering effects, such as haze, sandstorm, underwater, and remote sensing conditions, remains a fundamental yet challenging problem in computer vision. Existing methods either rely on a single prior, which is inherently insufficient to characterize diverse scattering degradations, or employ deep networks trained on synthetic data, which often suffer from limited generalization to real-world scenarios.
In this paper, we propose Spatial and Frequency Priors (SFP) for real-world scene recovery under scattering-induced degradations.
In the spatial domain, we observe that the inverse of a scattering-degraded image reveals a projection along its spectral direction that correlates with the underlying scene transmission. Based on this observation, a spatial prior is formulated to estimate the transmission map, enabling effective recovery of scene radiance under scattering effects.
In the frequency domain, we design an adaptive frequency enhancement strategy guided by two novel priors. The first prior assumes that the mean intensity of the direct current (DC) components across channels in degraded images approximates that of the corresponding clear images. 
The second prior is based on the observation that, in clear images, low radial frequencies within a narrow band contribute only a small proportion of the overall spectrum. These priors enable targeted compensation for scattering-induced attenuation across different frequency bands. 
Finally, a weighted fusion of the spatial and frequency domain results is performed to obtain the final recovered image.
Extensive experiments on diverse real-world scattering-degraded scenarios verify that our SFP achieves superior performance and strong generalization capability compared to state-of-the-art methods.

\end{abstract}


\section{Introduction}
In scattering-affected imaging scenarios (e.g., haze, sandstorms, underwater, etc), light interacting with particles in the medium is attenuated and deviated from its original propagation path, introducing an additive veiling component while reducing the direct transmission of scene radiance. As a result, the captured images suffer from significant quality degradation, manifested as reduced contrast, color distortion, and the loss of fine structural details. Such scattering-induced degradations not only impair visual perception but also substantially degrade the performance of downstream computer vision tasks, such as object detection, segmentation, and recognition. Therefore, scene recovery, which aims to recover clear and visually faithful images from scattering-degraded observations, is of fundamental importance for improving the reliability of vision-based systems in real-world applications.


Existing methods can be generally categorized into two groups: prior-based methods and learning-based methods. Prior-based methods~\cite{he2011single,zhu2015a,li2016underwater,zhang2017fast,ju2019idgcp,berman2020single,ju2021idrlp,zhuang2022underwater,ling2023single,wu2026image,liu2026ihdcp}, such as DCP~\cite{he2011single}, GCP~\cite{ju2019idgcp}, MRP~\cite{zhang2017fast}, HLP~\cite{berman2020single}, HLRP~\cite{zhuang2022underwater}, SLP~\cite{ling2023single}, PNBP~\cite{wu2026image}, and IHDCP~\cite{liu2026ihdcp}, typically rely on physical imaging models and exploit statistical priors to estimate scene transmission. However, these priors are specifically designed for a single task and are ineffective when dealing with diverse scattering degradations.
To address this limitation, several generalized priors have been proposed, such as GDCP~\cite{peng2018generalization}, ROP~\cite{liu2021rank}, ROP+~\cite{liu2023rank}, and ALSP~\cite{he2025alsp}, aiming to improve scene recovery across multiple degradation scenarios. However, their performance across diverse real-world scenarios remains limited, primarily because single-domain priors lack sufficient adaptability to handle the heterogeneous characteristics of scattering-induced degradations.


\begin{figure}[!t]
 \setlength{\abovecaptionskip}{2pt}   
    \setlength{\belowcaptionskip}{-15pt}  
    \centering    
    \includegraphics[width=3.5in]{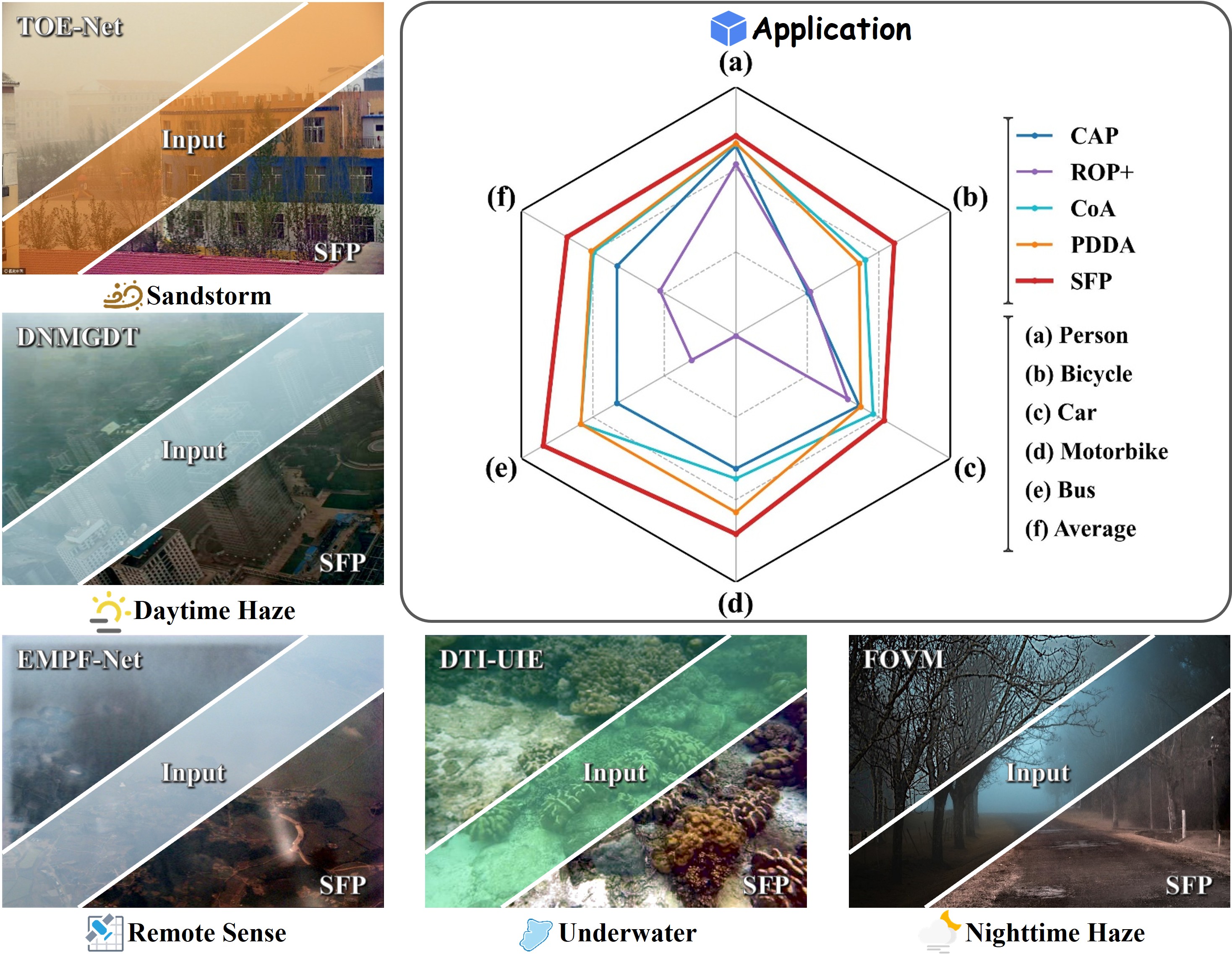}
    \caption{The recovered results of the proposed SFP under various scattering degradation scenarios and its performance comparison on the object detection task. Note that the data used for object detection are sourced from Table~\ref{tab:high_level}.}
    \label{fig:performance}
\end{figure}
Benefiting from recent advances in deep learning, a variety of neural networks~\cite{cai,valanarasu2022transweather,gao2023let,wen2023encoder,feng2024advancing,cong2024semi,lu2024aosrnet,su2025real,fu2025iterative,ning2025mabdt,liu2024real,li2025low,wen2025multi,cui2025adair,Chen_2025_CVPR,ma2025coa,zhang2026pdda,wang2026sea,lin2026downstream,gui2026brightness,cui2026starir} have been developed to address scattering-induced image degradations. Early learning-based methods are typically designed for specific scenarios. For instance, DehazeNet~\cite{cai}, KA-Net~\cite{feng2024advancing}, IPC-Dehaze~\cite{fu2025iterative}, CoA~\cite{ma2025coa}, SFSNiD~\cite{cong2024semi}, PDDA~\cite{zhang2026pdda}, and SFN~\cite{gui2026brightness} primarily focus on daytime or nighttime haze removal. TOE-Net~\cite{gao2023let} is intended for haze and sandstorm conditions. SEA-PACE~\cite{wang2026sea} and DTI-UIE~\cite{lin2026downstream} are designed for underwater scenarios. EMPF-Net~\cite{wen2023encoder} and MABDT~\cite{ning2025mabdt} aim to recover hazy scenes under remote sensing conditions.
More recently, several methods, such as AOSR-Net~\cite{lu2024aosrnet}, ERA-Net~\cite{liu2024real}, MPMF-Net~\cite{wen2025multi}, IDB~\cite{li2025low}, UniRestore~\cite{Chen_2025_CVPR}, AdaIR~\cite{cui2025adair}, and StarIR~\cite{cui2026starir}, have been proposed for image restoration, aiming to handle multiple types of scattering-induced degradations. Despite their promising performance, these learning-based approaches typically require substantial computational resources. More importantly, their effectiveness often depends on large-scale synthetic training data, which may not accurately reflect the complex and heterogeneous nature of real-world scattering-induced degradations, thereby limiting their generalization capability in practical scenarios.

In order to adapt to diverse real-world scattering-induced degradations without relying on large-scale training data or heavy computational resources, we propose a novel framework, called \textbf{S}patial and \textbf{F}requency \textbf{P}riors (SFP), for real-world scattering-degraded scene recovery. In the spatial domain, we introduce a new prior to estimate the transmission map by projecting the inverted input image along its spectral direction, followed by scene restoration via inversion of the atmospheric scattering model (ASM). In the frequency domain, we develop two priors based on the statistical characteristics of low-frequency components in clear and degraded images, which are used to construct an adaptive mask for frequency-domain enhancement. Finally, a weighted fusion strategy is employed to integrate complementary information from the spatial-domain restoration, frequency-domain enhancement, and the input image, producing the final recovered result.
Compared with existing prior- and learning-based methods, our SFP adapts well to diverse real-world degradations and achieves promising results, as illustrated in Fig.~\ref{fig:performance}. 

Overall, our main contributions are summarized as follows:
\begin{itemize}
    \item We propose a novel framework, termed Spatial and Frequency Priors (SFP), for real-world scene recovery under scattering-induced degradations. Unlike previous methods that rely on a single prior or domain, SFP jointly exploits complementary priors in both spatial and frequency domains to improve scene recovery performance.
    \item We propose a spatial domain prior and two frequency-domain priors. The spatial-domain prior leverages spectral properties to project the inverted degraded image onto the spectral direction, thereby approximating the scene transmission. The frequency-domain priors leverage the statistical characteristics of direct current (DC) components and low radial frequencies to construct an adaptive mask for degradation-aware enhancement.
    \item Experiments on real-world datasets demonstrate that SFP achieves superior and consistent performance across a variety of challenging scattering-induced scenarios, including haze, sandstorms, underwater environments, and remote sensing scenarios. Moreover, SFP generalizes well to complex real-world nighttime hazy scenes with coupled degradations and also improves the performance of downstream high-level vision tasks.
\end{itemize}

\section{Related Work}

\textbf{Prior-based Methods.} 
Based on physical imaging models, numerous priors have been proposed to solve the ill-posed inverse problem. The most representative one is the widely recognized DCP~\cite{he2011single}, assuming that in most non-sky regions, at least one color channel exhibits near-zero intensities, which allows the transmission to be estimated. Subsequently, CAP~\cite{zhu2015a} assumes that pixel brightness and saturation vary significantly with haze concentration and further constructs a linear model for scene depth estimation. GCP~\cite{ju2019idgcp} modifies gamma correction as a preprocessing step based on the observation that inverted low-light images resemble hazy images. RLP~\cite{ju2021idrlp} exhibits a quasi-linear relationship between the average pixel intensities in hazy and corresponding haze-free regions. SLP~\cite{ling2023single} derives a linear relationship between the saturation component and the reciprocal of the brightness component. IHDCP~\cite{liu2026ihdcp} models scene transmission from the inverted haze density map via a pixel-wise gamma correction. 
Despite providing effective dehazing performance, the aforementioned priors are limited in other degradation scenarios, such as nighttime haze. To achieve nighttime dehazing, Zhang et al.~\cite{zhang2017fast} propose a new prior named MRP to estimate the varying ambient illumination, based on the observation that daytime haze-free image patches often have high-intensity pixels in each color channel. In addition, variation priors, such as UVRM~\cite{liu2023multi}, VNDHR~\cite{liu2025vndhr} and FOVM~\cite{liu2026fovm}, are specifically designed for nighttime haze removal to regularize the model. For underwater conditions, HLRP~\cite{zhuang2022underwater} introduces a variational Retinex model to enhance degraded images. These approaches achieve good results for their respective tasks; however, they fail to generalize to other degradation scenarios. To simultaneously overcome multiple challenging degradations, 
GDCP~\cite{peng2018generalization} estimates scene transmission via the scene ambient light differential derived from depth-dependent color variation, enabling effective restoration of hazy, sandstorm, and underwater images. Subsequently, Liu et al.~\cite{liu2021rank,liu2023rank} propose a rank-one prior (ROP) based on an intensity projection strategy to tackle daytime haze, sandstorms, and underwater conditions. More recently, ALSP~\cite{he2025alsp} introduces the ambient light similarity metric and leverages the relationship between transmission and this metric for scene recovery.
Nevertheless, their scene recovery performance in real-world scenarios remains limited, as they primarily consider single-domain degradations that are insufficient to capture the complex and heterogeneous nature of scattering-induced degradations. In contrast, our SFP jointly exploits spatial and frequency priors for effective scene recovery under diverse degradations.

\textbf{Learning-based Methods.}
Leveraging the powerful nonlinear modeling capability of deep learning, several network architectures~\cite{cai,ren2016single,wen2023encoder,feng2024advancing,cong2024semi,su2025real,fu2025iterative,ning2025mabdt,li2025low,ma2025coa,zhang2026pdda,gui2026brightness,lin2026downstream,wang2026sea} are designed for a single degradation task. 
For instance, DehazeNet~\cite{cai} and MSCNN~\cite{ren2016single} learn to estimate transmission maps directly from hazy inputs. 
Furthermore, KA-Net~\cite{feng2024advancing}, IPC-Dehaze~\cite{fu2025iterative}, and PDDA~\cite{zhang2026pdda} aim to improve generalization for real-world dehazing. However, these methods often struggle with nighttime scenes due to the significant domain gap between daytime and nighttime haze.
To achieve challenging nighttime dehazing, SFSNiD~\cite{cong2024semi} and SFN~\cite{gui2026brightness} are specifically designed for nighttime dehazing.
In addition, EMPF-Net~\cite{wen2023encoder} and MABDT~\cite{ning2025mabdt} focus on remote sensing haze removal. For underwater degraded scenes, SEA-PACE~\cite{wang2026sea} exploits semi-supervised learning with uncertainty modeling for underwater image enhancement. DTI-UIE~\cite{lin2026downstream} incorporates downstream task guidance to improve detail reconstruction and recognition performance.
Recently, several image restoration networks~\cite{gao2023let,lu2024aosrnet,liu2024real,li2025low,wen2025multi,cui2025adair,cui2025adair,Chen_2025_CVPR,cui2026starir} have been proposed to handle multiple degradations. 
TOE-Net~\cite{gao2023let} jointly addresses haze and sandstorm degradations by modeling channel-wise correlations through an MLP-based module, enabling unified enhancement of low-visibility images under different scattering conditions.
AOSR-Net~\cite{lu2024aosrnet} integrates multiple enhancement modules, including gamma correction and linear stretching, to improve visibility and restore details across diverse low-visibility scenarios.
Wen et al.~\cite{wen2025multi} propose MPMF-Net, which leverages multi-axis prompt learning and multi-dimension feature interaction to restore weather-degraded images. 
AdaIR~\cite{cui2025adair} mines both spatial and frequency domain information to address different types of degradations.
Despite their impressive performance on simulated degraded images, learning-based approaches typically exhibit limited generalization to real-world scenes, primarily due to domain discrepancies. Moreover, data-driven methods require large-scale datasets for training and considerable computational resources.

\begin{figure}[!t]
 \setlength{\abovecaptionskip}{2pt}   
    \setlength{\belowcaptionskip}{-13pt}  
    \centering    
    \includegraphics[width=3.5in]{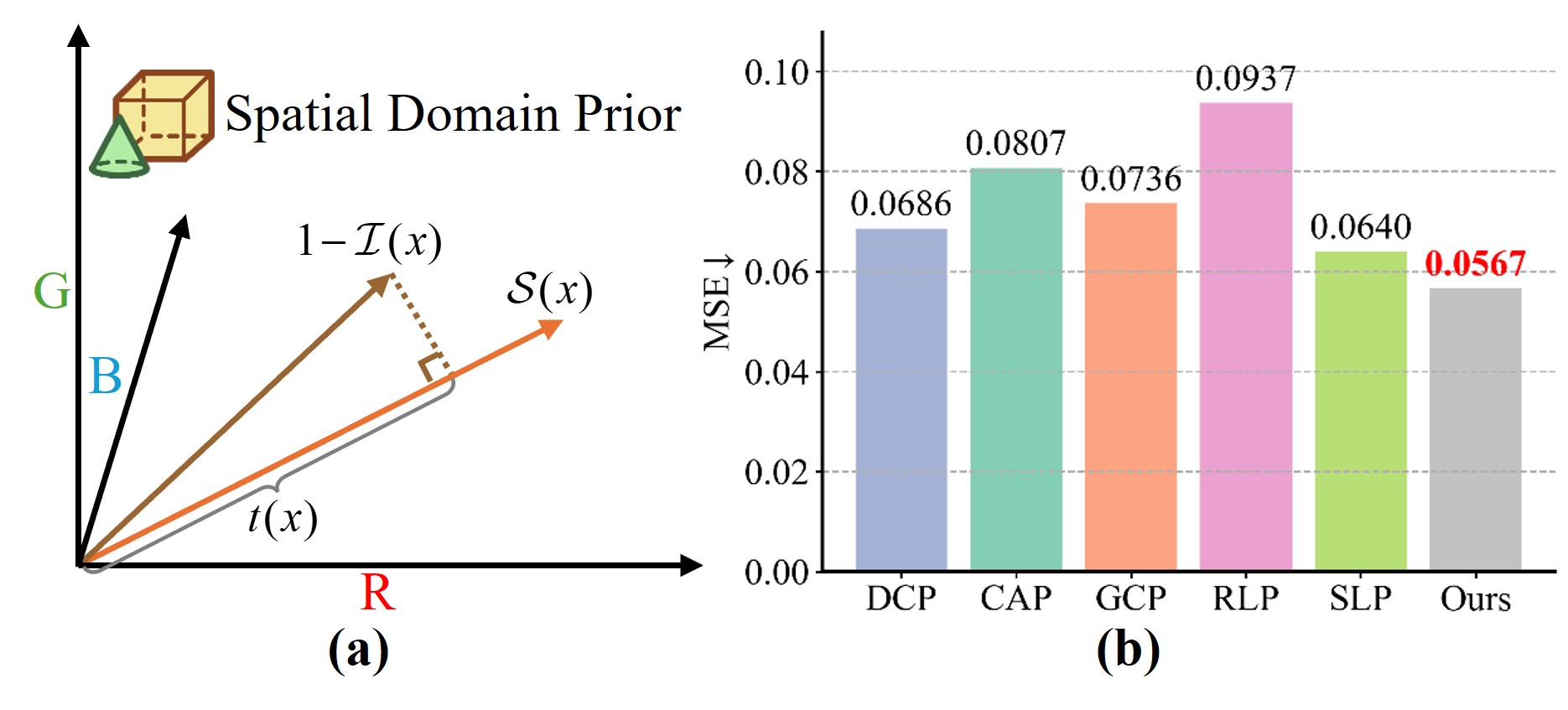}
    \caption{ Illustration of our spatial-domain prior. (a) Schematic diagram of approximating the transmission map $t(x)$ by projecting $1-\mathcal{I}(x)$ onto the spectral direction $\mathcal{S}(x)$. (b) Comparison of MSE deviations between estimated and ground-truth transmissions for various priors on the Haze4K test set. }
    \label{fig:spatial_prior}
\end{figure}

\section{Spatial and Frequency Priors (SFP)}
In this section, we introduce the proposed Spatial and Frequency Priors (SFP), including a spatial-domain prior and two frequency-domain priors.

\subsection{Spatial-Domain Prior (SDP)}
The widely used atmospheric scattering model (ASM)~\cite{narasimhan2002vision} for describing the scene scattering process is expressed as:
\begin{equation}
\mathcal{I}(x) = \mathcal{J}(x) \circ t(x) + \mathcal{A} \circ (1 - t(x)),
\label{Eq:asm}
\end{equation}
where $\circ$ denotes the Hadamard product. $\mathcal{I}(x)$  and  $\mathcal{J}(x)$ represent the degraded image and the corresponding clear image, respectively. $t(x)$ stands for the medium transmission, and $\mathcal{A}$ indicates the global atmospheric light. 
Owing to the ill-posed nature of model inversion, a novel spatial-domain prior that accounts for spectral properties is proposed to estimate scene transmission.

The spectral direction represents color variations within an image and can be approximately modeled as the gradients of the three channels, which capture the rate of intensity change across spatial locations. Accordingly, the spectral direction $\mathcal{S}(x)$ is defined as follows:
\begin{equation}
\mathcal{S}(x) =
\frac{
\displaystyle \sum_{x' \in \Omega(x)}
\dfrac{
\big(\nabla \mathcal{I}_{\mathcal{R}}(x'),\, \nabla \mathcal{I}_{\mathcal{G}}(x'),\, \nabla \mathcal{I}_{\mathcal{B}}(x')\big)
}{
\big\| \nabla \mathcal{I}(x') \big\|}
}{
|\Omega(x)|},
\label{spectral_direction}
\end{equation}
where $\mathcal{R}$, $\mathcal{G}$, and $\mathcal{B}$ denote the color channels, ${\Omega (x)}$ is the local patch centered at $x$, and $\left\| {\cdot} \right\|$ is the modulus operation.

We observe that projecting the inverted degraded image $1 - \mathcal{I}(x)$ onto the spectral direction $\mathcal{S}(x)$ provides a close approximation of the scene transmission, as illustrated in Fig.~\ref{fig:spatial_prior}(a). Intuitively, $1 - \mathcal{I}(x)$ serves as an indicator of haze density by quantifying a pixel's deviation from the atmospheric light. This deviation is monotonically related to the scene transmission $t(x)$, decreasing as haze density increases. By projecting $1 - \mathcal{I}(x)$ onto the spectral direction $\mathcal{S}(x)$ of the degraded image, the degradation-related component can be effectively extracted, yielding an accurate approximation of the scene transmission.



\begin{figure}[!t]
 \setlength{\abovecaptionskip}{2pt}   
    \setlength{\belowcaptionskip}{-13pt}  
    \centering    
    \includegraphics[width=3.5in]{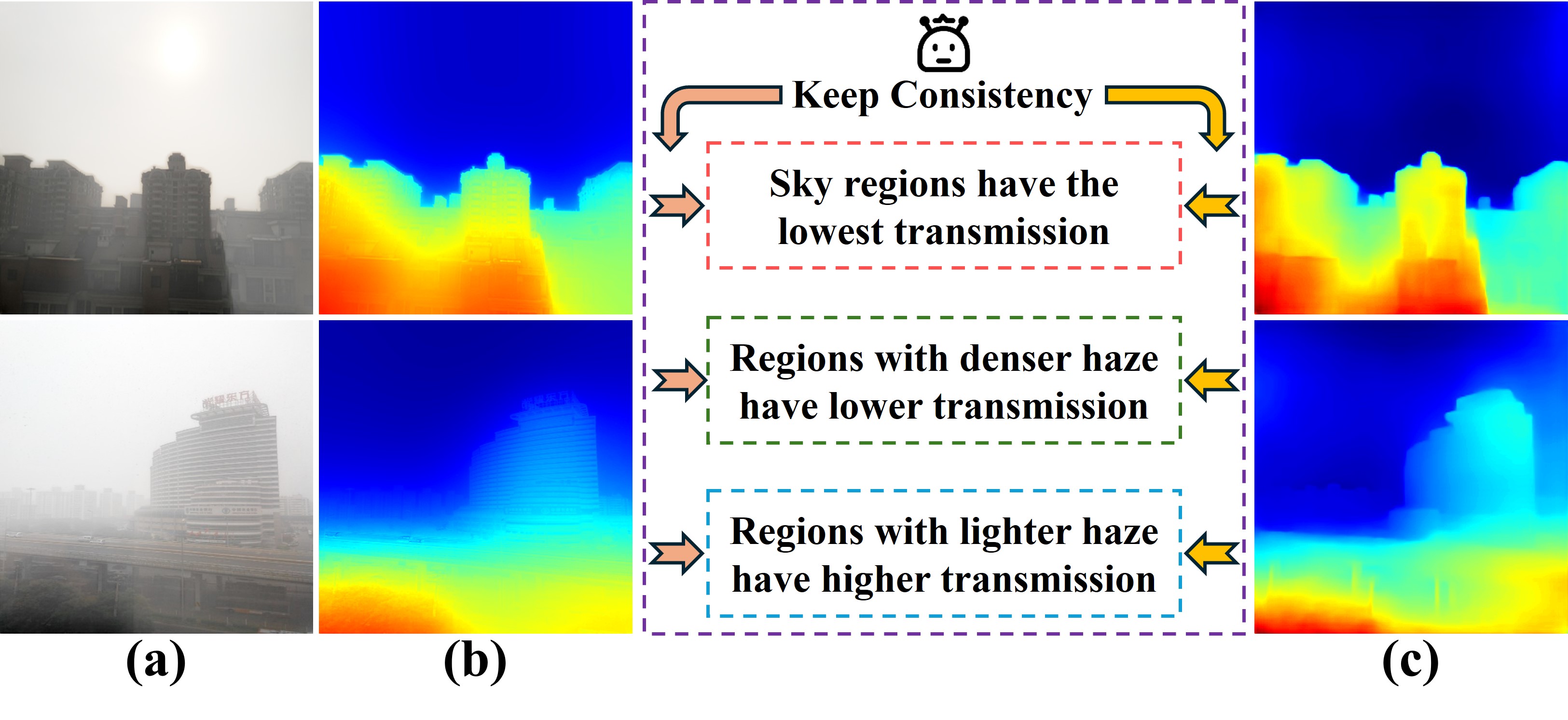}
    \caption{The projection of the inverted degraded image $1-\mathcal{I}(x)$ onto the spectral direction $\mathcal{S}(x)$ closely resembles the ground-truth scene transmission. (a) Degraded images. (b) Transmission maps estimated by our spatial-domain prior. (c) Ground-truth transmission maps. }
    \label{fig:similar_gt}
\end{figure}

To verify this finding, we compute the mean squared error (MSE) to quantify the discrepancy between the estimated and ground-truth transmissions on the Haze4K test set~\cite{liu2021synthetic}, with the statistical results shown in Fig.~\ref{fig:spatial_prior}(b). The proposed spatial-domain prior achieves superior performance, yielding estimates with smaller deviations from the ground truth compared to several famous priors, e.g. DCP~\cite{he2011single}, CAP~\cite{zhu2015a}, GCP~\cite{ju2019idgcp}, RLP~\cite{ju2021idrlp}, and SLP~\cite{ling2023single}. 
Fig.~\ref{fig:similar_gt} shows that our spatial-domain prior yields transmission maps which remain consistent with the ground truth across regions of varying haze concentrations.






\subsection{Frequency-Domain Priors (FDP)}
Scattering-induced degradations, such as haze, underwater, and sandstorm conditions, primarily affect the low-frequency components of images. Therefore, we attempt to conduct extensive statistical analyses to characterize the differences and correlations between the low-frequency components of degraded and clear images. Based on these observations, we propose two frequency-domain priors to enable adaptive frequency enhancement of degraded images.



The first frequency-domain prior is motivated by the observation that, in the frequency domain, the DC component of each channel $\mathcal{F}(\mathcal{E}_{\mathcal{C}})(0)$ in a clear image closely approximates the mean DC value across the three channels of its corresponding degraded image. Mathematically, this prior can be expressed as:
\begin{equation}
\left\{ {\begin{array}{l}
{{\cal F}({{\cal E}_{\cal C}})(0) \approx \dfrac{1}{3}\sum\limits_{{\cal C} \in \{ {\cal R},{\cal G},{\cal B}\} } {\cal F} ({{\cal I}_{\cal C}})(0)}\\
{{\cal F}({{\cal I}_{\cal C}})(0) = \sum\limits_{x \in \Omega } {{{\cal I}_{\cal C}}} (x)}\\
{{\cal F}({{\cal E}_{\cal C}})(0) = \sum\limits_{x \in \Omega } {{{\cal E}_{\cal C}}} (x)}
\end{array}} \right.
\end{equation}
where $\mathcal{E}_{\mathcal{C}}$ denotes the clear image in the $\mathcal{C}$-th channel, $\mathcal{F}(\cdot)$ represents the 2D Fast Fourier Transform (FFT), and $|\Omega| = \mathcal{H} \times \mathcal{V}$, with $\mathcal{H}$ and $\mathcal{V}$ denoting the height and width of the image, respectively.

\begin{figure}[!t]
 \setlength{\abovecaptionskip}{2pt}   
    \setlength{\belowcaptionskip}{-13pt}  
    \centering    
    \includegraphics[width=3.5in]{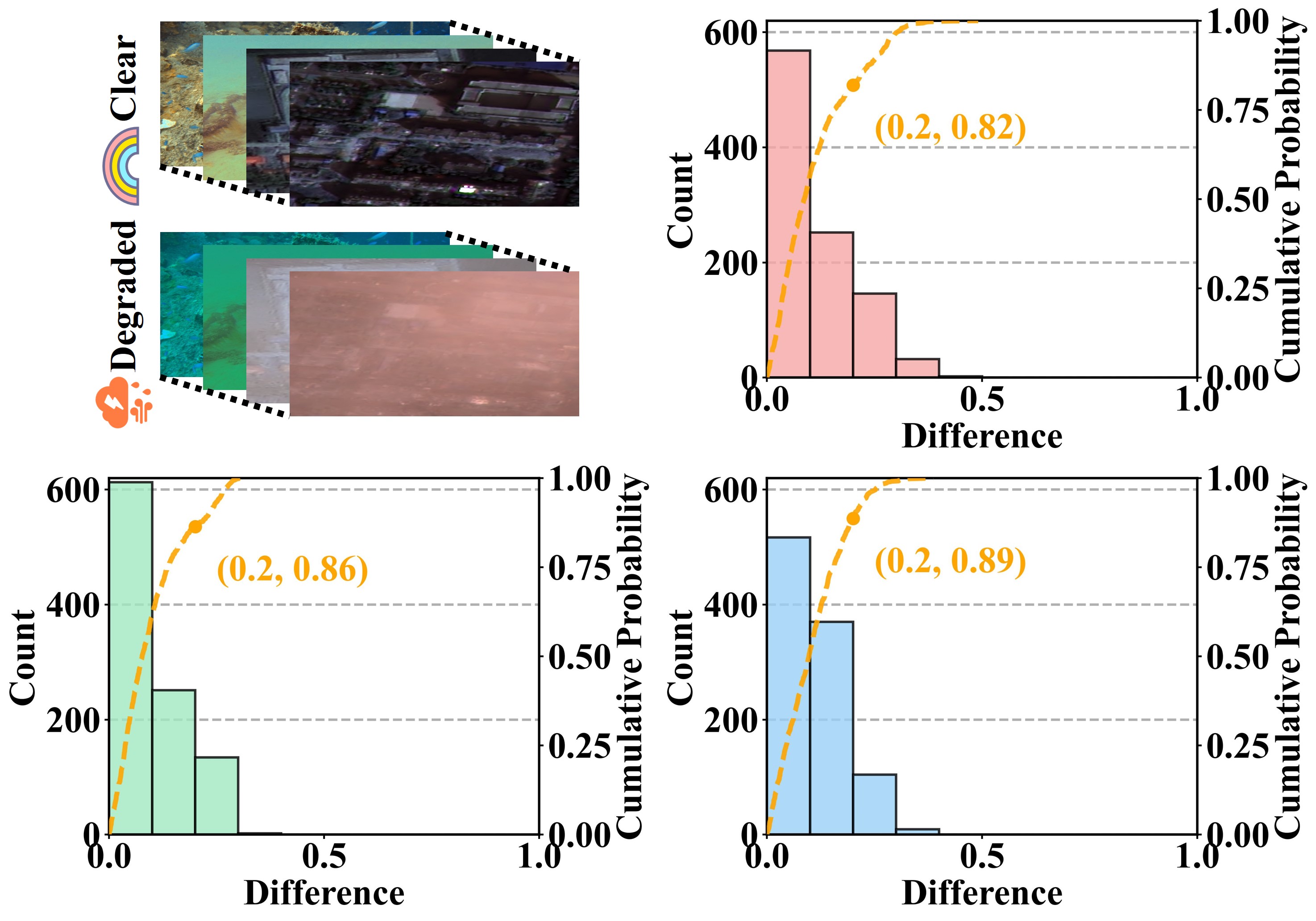}
    \caption{Statistical distributions of the absolute differences between the normalized DC component (DC$\in \left[ {0,1} \right]$) of each channel (e.g. $\mathcal{R}$, $\mathcal{G}$, $\mathcal{B}$) in clear images and the mean DC value across the three channels of their corresponding degraded images, computed on 1000 degraded-clean image pairs. The yellow dashed curves (cumulative distributions) show that over 80\% of image pairs have absolute differences below 0.2 in all three channels.}
    \label{fig:fdp_dc}
\end{figure}

To validate this prior, we conducted a statistical analysis on 1000 randomly selected degraded and clear image pairs from the UIEB~\cite{li2019underwater} and RRSHID~\cite{zhu2025real} datasets. For each pair, we calculate the absolute difference between the DC component of each channel in the clear image and the mean DC value across the three channels of the corresponding degraded image. In Fig.~\ref{fig:fdp_dc}, the majority of image pairs (over 80\%) exhibit absolute differences below 0.2, with some differences even approaching zero, indicating that the mean DC value of the degraded image across its three RGB channels is closely aligned with the DC component of each channel in the clear image.



\begin{figure*}[!t]
 \setlength{\abovecaptionskip}{2pt}   
    \setlength{\belowcaptionskip}{-13pt}  
    \centering    
    \includegraphics[width=7.15in]{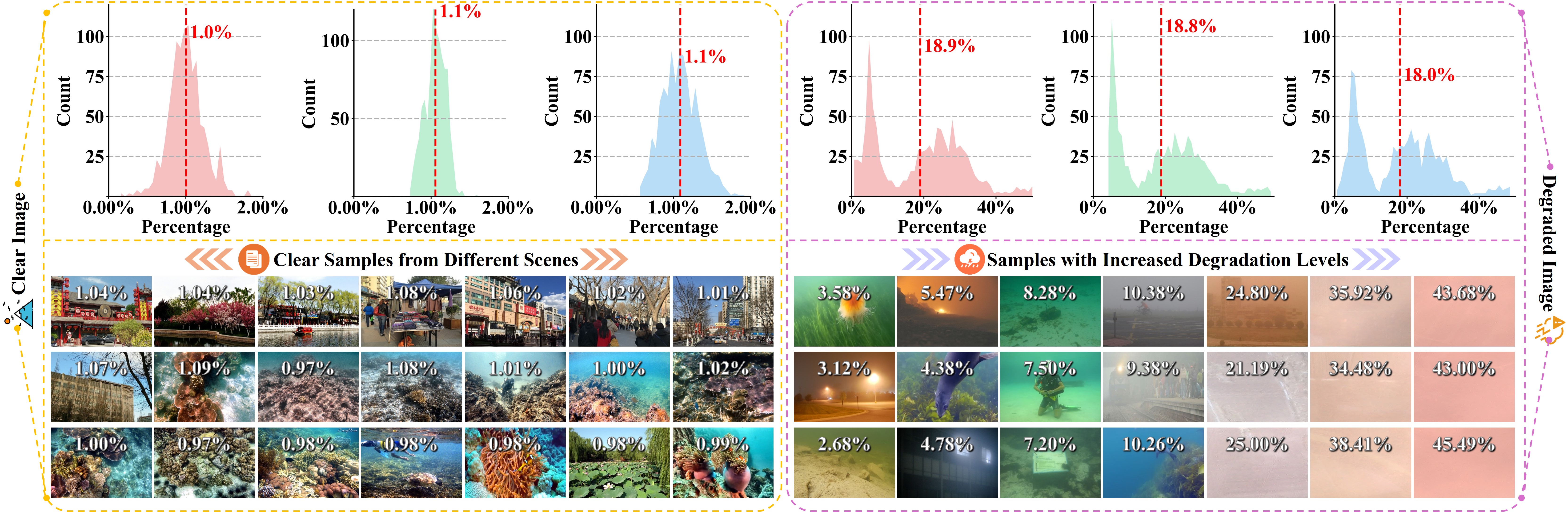}
    \caption{Statistical distributions of the percentages of low radial frequency components ($<$0.001) in the $\mathcal{R}$, $\mathcal{G}$, and $\mathcal{B}$ channels for 1000 clear  and 1000 degraded images. The horizontal axis represents the percentage of spectral magnitudes ($<$0.001) in the total spectral magnitude, while the vertical axis indicates the number of images. The red dashed lines denote the mean values for each channel. The bottom row presents examples of clear and degraded images along with their corresponding low radial frequency percentages ($<$0.001). }
    \label{fig:fdp_radial}
\end{figure*}

The second frequency-domain prior is based on the observation that, for clear images, the proportion of low radial frequencies (i.e., $\rho(x) < 0.001$) within the total spectral magnitude is approximately $1\%$, while higher proportions indicate more severe degradation. 
Let $\Omega_{\text{low}} = {x \mid \rho(x) < 0.001}$ denote the extremely low-frequency region. Accordingly, this prior can be formulated as:
\begin{equation}
\Phi  = \dfrac{{\int_{{\Omega _{low}}} | {\cal F}({{\cal I}_{\cal C}})(x)|dx}}{{\int {|{\cal F}({{\cal I}_{\cal C}})(x)|dx} }} \approx 1\%,
\end{equation}
where $\rho (x)$ is the normalized radial frequency, denoted as:
\begin{equation}
 \rho(x) = \dfrac{\sqrt{(u - \mathcal{H}/2)^2 + (v - \mathcal{V}/2)^2}}{\sqrt{(\mathcal{H}/2)^2 + (\mathcal{V}/2)^2}}, where~x=(u,v)
\end{equation}
where $(u,v)$ is the frequency coordinate.

To confirm the validity of this prior, 
we randomly collect 1000 high-quality clear images and 1000 degraded images, and perform statistical analyses on the proportion of low-frequency magnitudes (i.e., $\rho(x) < 0.001$) within the total spectral magnitude for each RGB channel. As shown in Fig.~\ref{fig:fdp_radial}, 
the distributions for clear images are concentrated around $1\%$, whereas degraded images exhibit significantly higher proportions. Furthermore, visual examples annotated with their corresponding low-frequency proportions demonstrate a clear correlation: values for clear scenes cluster near $1\%$ and increase progressively with the severity of degradation.





\section{Scene Recovery Using SFP}
In this section, we present the scene recovery framework based on the proposed Spatial and Frequency Priors (SFP).
\subsection{Spatial-Domain Restoration}
Based on our proposed spatial-domain prior, we project the inverted degraded image $1-\mathcal{I}(x)$ onto the spectral direction $\mathcal{S}(x)$ to estimate the scene transmission:
\begin{equation}
t(x) = \dfrac{\sum\limits_{\mathcal{C}} \left\langle \mathcal{S}_{\mathcal{C}}(x), 1 - \mathcal{I}_{\mathcal{C}}(x) \right\rangle \circ \mathcal{S}_{\mathcal{C}}(x)}{3\|\mathcal{S}_{\mathcal{C}}(x)\|^2}, \mathcal{C} \in \{\mathcal{R}, \mathcal{G}, \mathcal{B}\}
\label{transmission}
\end{equation}
where $\left\langle { \cdot , \cdot } \right\rangle $ denotes the inner product and $\| \cdot \|^2$ is the  squared $\ell_2$ norm.

To estimate the atmospheric light, we select the lowest 0.1\% of the pixels in the transmission map and compute their average value in the observed image $\mathcal{I}$ as $\mathcal{A}$. Once $t$ and $\mathcal{A}$ are obtained, the recovered image in the spatial domain is derived:
\begin{equation}
\mathcal{J}(x) = {({\mathcal{I}(x) - \mathcal{A}}) \mathord{\left/
 {\vphantom {{\mathcal{I}(x) - \mathcal{A}} {\mathcal{GF}(max (t(x),{t_{min }}))}}} \right.
 \kern-\nulldelimiterspace} {\mathcal{GF}(max (t(x),{t_{min }}))}} + \mathcal{A},
 \label{asm_recovery}
\end{equation}
where $\mathcal{GF}$ denotes the guided filter~\cite{he2013}, and $t_{min }$ is a lower bound set to the average of the lowest $5\%$ of $t(x)$. 

\begin{figure}[!t]
 \setlength{\abovecaptionskip}{2pt}   
    \setlength{\belowcaptionskip}{-13pt}  
    \centering    
    \includegraphics[width=3.5in]{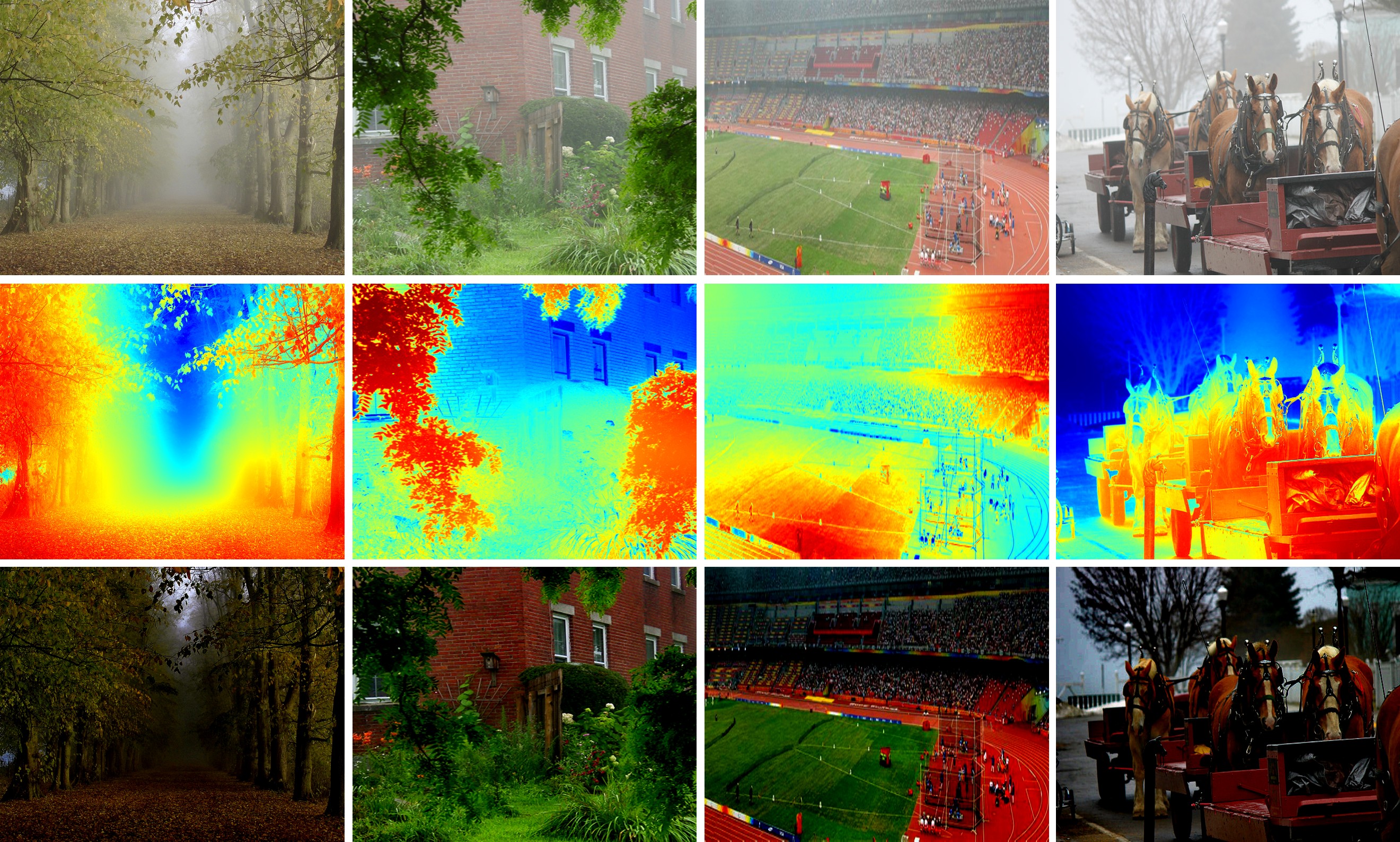}
    \caption{Scene recovery results on real-world hazy images using the spatial-domain prior. The first row shows the input images, and the second and third rows depict the estimated transmission maps and the corresponding recovery results, respectively. }
    \label{fig:sdp_effect}
\end{figure}

As depicted in Fig.~\ref{fig:sdp_effect}, the proposed SDP is able to effectively suppress the major degradation effects in real-world scenarios, leading to clearer structural visibility and significantly reduced haze. Nevertheless, some limitations can still be observed in the recovered results, particularly in terms of detail loss and diminished brightness.

\subsection{Frequency-Domain Enhancement}

To enhance the frequency components of degraded images, we design a simple yet effective adaptive mask derived from our proposed frequency-domain priors. Formally, the enhancement is expressed as follows:
\begin{equation}
\mathcal{E}_\mathcal{C}(x) = {\mathcal{F}^{ - 1}}(\mathcal{F}({\mathcal{I}_\mathcal{C}})(x) \circ {\mathcal{M}_\mathcal{C}}(x)),
\label{process}
\end{equation}
where $\mathcal{E}$ represents the enhanced clear result, $\mathcal{F}(\cdot)$ and $\mathcal{F}^{ - 1}(\cdot)$ denote the 2D FFT and its inverse, respectively. The adaptive mask $\mathcal{M}_\mathcal{C}(x)$ is defined as:
\begin{equation}
{\mathcal{M}_\mathcal{C}}(x) = {\alpha _\mathcal{C}} - {e^{( - {{({{\rho (x)} \mathord{\left/
 {\vphantom {{\rho (x)} {{\beta _\mathcal{C}}}}} \right.
 \kern-\nulldelimiterspace} {{\beta _\mathcal{C}}}})}^2})}},
 \label{eq:mask}
\end{equation}
where $\alpha _\mathcal{C}$ and $\beta _\mathcal{C}$ are two parameters that control the magnitude of frequency enhancement, which are adaptively determined by our two frequency-domain priors.

Based on the first frequency-domain prior, we define $\mu$ to represent the DC component:
\begin{equation}
\mathcal{F}({\mathcal{E}_\mathcal{C}})(0) = \dfrac{{\sum\limits_\mathcal{C} {\mathcal{F}({\mathcal{I}_\mathcal{C}})(0)} }}{3}=\mu, \mathcal{C} \in \{ \mathcal{R},\mathcal{G},\mathcal{B}\} ,
\label{mean_relationship}
\end{equation}

Then, by taking the 2D FFT of both sides of Eq.~(8), we can obtain:
\begin{equation}
{\cal F}\left( {{{\cal E}_{\cal C}}} \right)(x) = {\cal F}({{\cal I}_{\cal C}})(x) \circ {{\cal M}_{\cal C}}(x),
\label{process_variant}
\end{equation}

By setting $x=0$ in Eq.~(\ref{process_variant}) and combining it with Eq.~(\ref{mean_relationship}), we can derive:
\begin{equation}
\mathcal{F}({\mathcal{E}_\mathcal{C}})(0)=\mu=\mathcal{F}({\mathcal{I}_\mathcal{C}})(0) \circ {\mathcal{M}_\mathcal{C}}(0),
\label{process_mean}
\end{equation}

Given that the DC component corresponds to the zero radial frequency $\rho(0)=0$, Eq.~(\ref{eq:mask}) gives:
\begin{equation}
 \mathcal{M}_{\mathcal{C}}(0) = \alpha_{\mathcal{C}} - e^{0} = \alpha_{\mathcal{C}} - 1 ,
\label{rho_0}
\end{equation}

By substituting Eq.~(\ref{rho_0}) into Eq.~(\ref{process_mean}), we have:
\begin{equation}
\mu=\mathcal{F}({\mathcal{I}_\mathcal{C}})(0)\circ ({\alpha _\mathcal{C}} - 1),
\label{eq:miu}
\end{equation}

Furthermore, $\alpha _\mathcal{C}$ is determined by rearranging Eq.~(\ref{eq:miu}):
\begin{equation}
{\alpha _{\cal C}} = {\mu  \mathord{\left/
 {\vphantom {\mu  {{\cal F}({{\cal I}_{\cal C}})(0)}}} \right.
 \kern-\nulldelimiterspace} {{\cal F}({{\cal I}_{\cal C}})(0)}} + 1,
 \label{alpha}
\end{equation}

\begin{figure}[!t]
 \setlength{\abovecaptionskip}{2pt}   
    \setlength{\belowcaptionskip}{-13pt}  
    \centering    
    \includegraphics[width=3.5in]{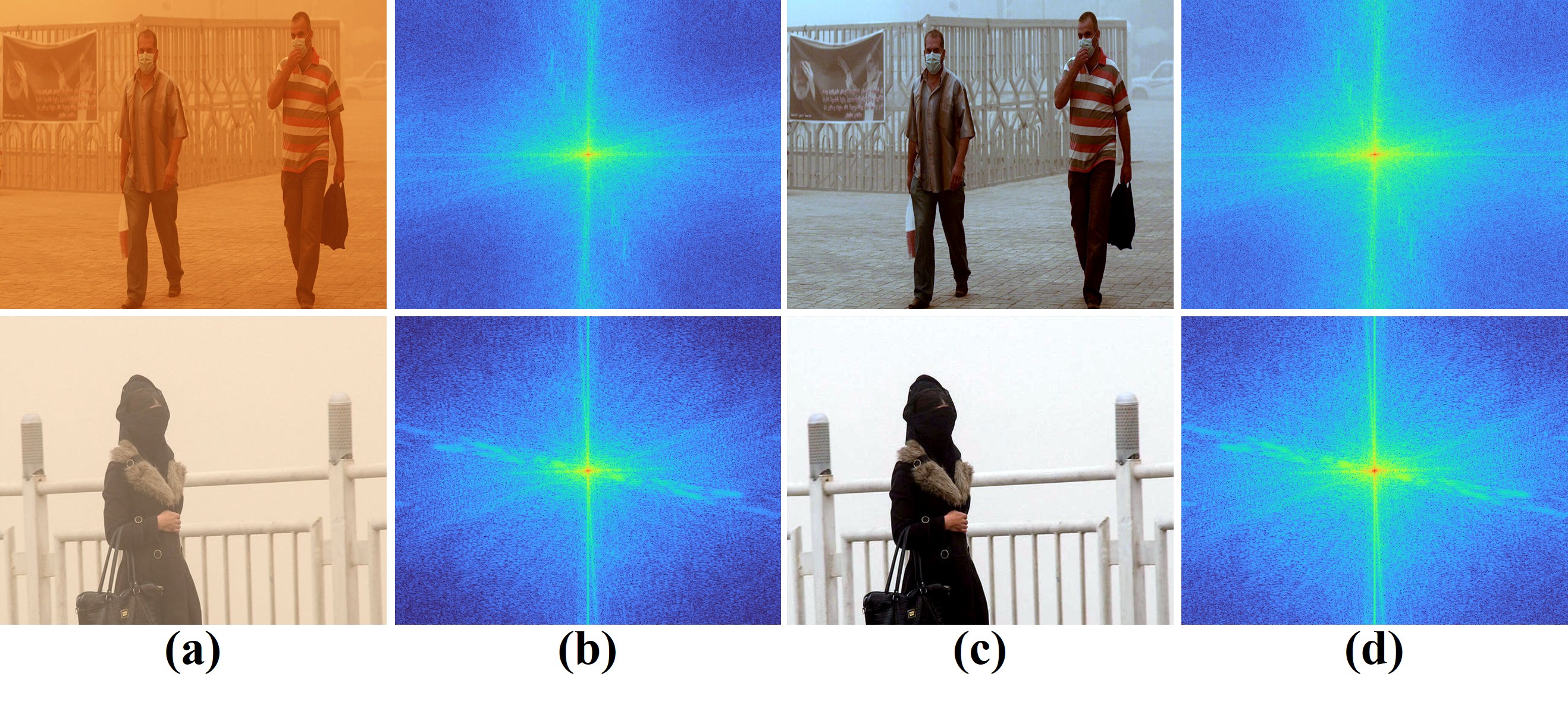}
    \caption{Scene recovery results on real-world sandstorm images using frequency-domain priors. (a) and (c) denote the input images and their corresponding recovery results, respectively. (b) and (d) illustrate the corresponding amplitude spectra in the frequency domain. }
    \label{fig:frequency_adjust}
\end{figure}
Based on the second frequency-domain prior, we formulate the following energy function to estimate $\beta_{\mathcal{C}}$, with the objective of bringing the percentage of low radial frequencies ($\rho(x) < 0.001$) close to 1\%:
\begin{equation}
\mathop {\arg \min }\limits_{{\beta _\mathcal{C}}} \left\{ {|\Phi ({\cal P}({\beta _\mathcal{C}})) - 1\%|} \right\},
\label{beta}
\end{equation}
where $\Phi$ represents a function that calculates the percentage of low radial frequencies ($\rho(x) < 0.001$) and $\mathcal{P}$ denotes the enhanced result associated with the parameter ${\beta _\mathcal{C}}$. The above one-dimensional optimization problem can be efficiently solved using the fminbnd method.

As illustrated in Fig.~\ref{fig:frequency_adjust}, the proposed frequency-domain priors effectively regulate scattering-induced brightness and color distortions through frequency enhancement. The corresponding amplitude spectra of the recovered results clearly indicate enriched high-frequency details after enhancement.


\begin{figure}[!t]
 \setlength{\abovecaptionskip}{2pt}   
    \setlength{\belowcaptionskip}{-13pt}  
    \centering    
    \includegraphics[width=3.5in]{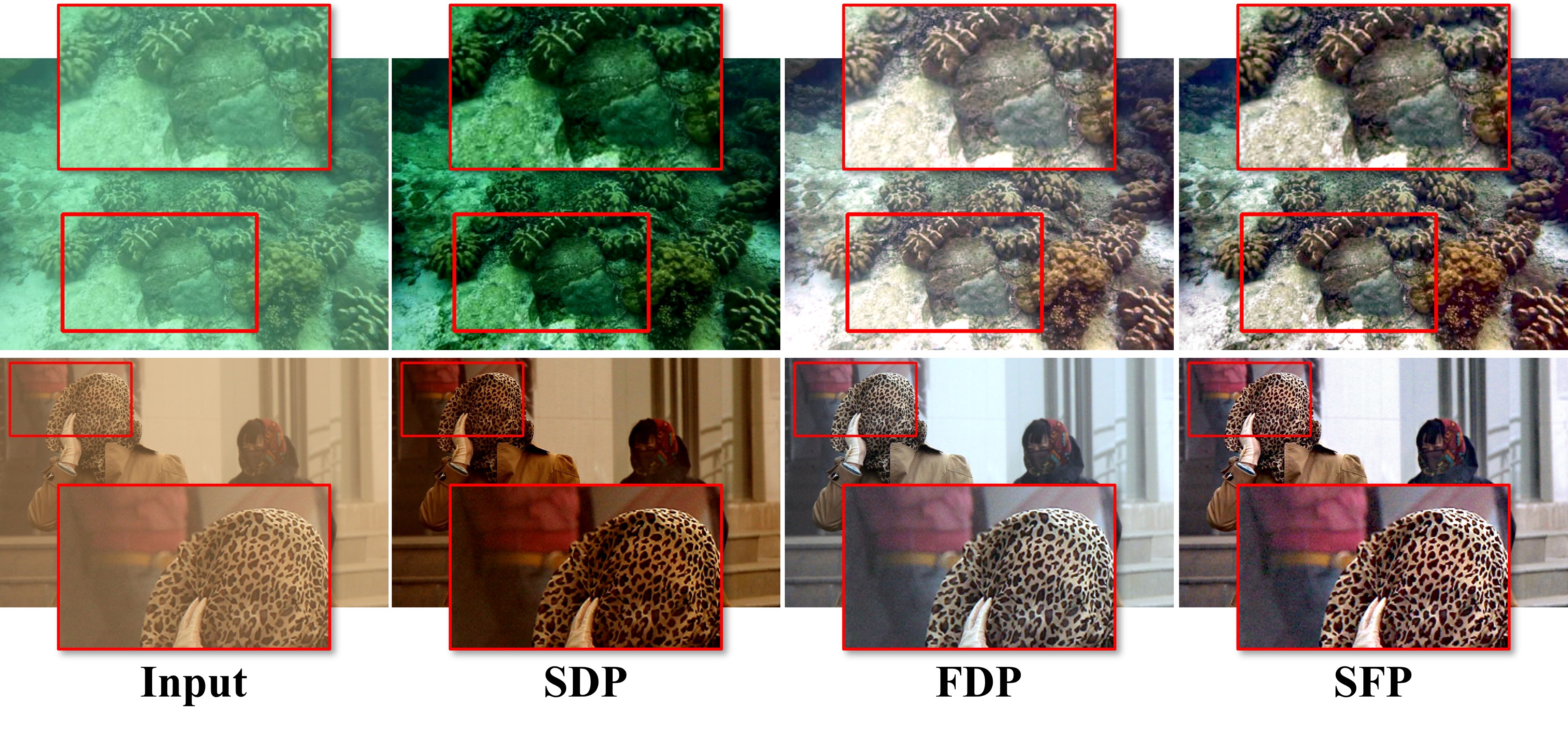}
    \caption{Effectiveness of the weight fusion strategy for scene recovery.  }
    \label{fig:fusion_effect}
\end{figure}

To fully integrate the complementary information from the spatial-domain restored image, the frequency-domain enhanced image, and the input degraded image, a simple weighted fusion strategy is introduced. Specifically, all three images are first converted into the Lab color space to separately mitigate color distortion and enhance brightness. Subsequently, a weighted fusion is employed in the $a$ and $b$ channels:
\begin{equation}
\mathcal{O}_{lab}^{a/b} = \sum\limits_{\mathcal{K} \in \{ \mathcal{I}, \mathcal{J},\mathcal{E} \}} \mathcal{W}_{\mathcal{K}}^{a/b} \circ \mathcal{K}_{lab}^{a/b},
\label{Eq:fusion}
\end{equation}
where $\mathcal{W}_\mathcal{I}^{a/b}$, $\mathcal{W}_\mathcal{J}^{a/b}$, and $\mathcal{W}_\mathcal{E}^{a/b}$ are the weights of $\mathcal{I}_{lab}^{a/b}$, $\mathcal{J}_{lab}^{a/b}$, and $\mathcal{E}_{lab}^{a/b}$, respectively. 
Leveraging the fundamental property that color-balanced images have near-zero means in the $a$ and $b$ channels of the Lab color space, we design the weights to fully utilize color-neutral information:
\begin{equation}
\mathcal{W}_{\mathcal{K}}^{a/b} = \frac{e^{-\left| \mathcal{M}_{\mathcal{K}}^{a/b} \right|}}{\sum\limits_{\mathcal{K} \in \{ \mathcal{I}, \mathcal{J}, \mathcal{E} \}} e^{-\left| \mathcal{M}_{\mathcal{K}}^{a/b} \right|}}, \quad \mathcal{K} \in \{ \mathcal{I}, \mathcal{J}, \mathcal{E} \}
\end{equation}
where ${M_{\cal I}^{a/b}}$, ${M_{\cal J}^{a/b}}$, and ${M_{\cal E}^{a/b}}$ are the mean values of the $a/b$ color channels of $\mathcal{I}$, $\mathcal{J}$ and $\mathcal{E}$, respectively.

For the $l$ channel, a discrete wavelet transform (DWT) is applied, in which the low-frequency component $\mathcal{J}^{l}_{low}$ is retained and the high-frequency components are fused using a maximum-value strategy. Finally, $\mathcal{O}_{lab}$ is transformed back into the RGB color space to obtain the fused image.

As illustrated in Fig.~\ref{fig:fusion_effect}, SDP primarily suppresses scattering-induced degradations and preserves structural details, while FDP enhances frequency components to correct color distortions and improve visual fidelity. These two priors provide complementary benefits, one focusing on spatial restoration and the other on frequency enhancement, which motivates their integration within the proposed fusion framework. As shown, by combining both cues, the SFP achieves more visually pleasing results than using either prior alone, with clearer structures and more natural colors.

\begin{algorithm}[t]
\caption{Scene Recovery Using SFP}
\label{procedure}
\KwIn{Degraded image $\mathcal{I}(x)$.}
\KwOut{Clear image $\mathcal{O}(x)$.}
Estimate transmission $t(x)$ using Eq.~(\ref{transmission});\\
Obtain the spatial restored image using Eq.~(\ref{asm_recovery});\\
Compute $\alpha_\mathcal{C}$ using Eq.~(\ref{alpha});\\
Compute $\beta_\mathcal{C}$ using Eq.~(\ref{beta});\\
Obtain the frequency enhanced image using Eq.~(\ref{process});\\
Obtain the final result $\mathcal{O}(x)$ using Eq.~(\ref{Eq:fusion});\\
\Return $\mathcal{O}(x)$;

\end{algorithm}

\begin{table*}[t]
	\renewcommand\arraystretch{1.2}
    \setlength{\tabcolsep}{9pt}
	\centering
    \caption{Quantitative comparison of the proposed SFP and state-of-the-art methods on four real-world hazy image datasets.}
    \vspace{-0.2cm}
	\begin{tabular}{c|cc|cc|cc|cc|cc}
		\toprule
		\multirow{2}{*}{Methods}
		& \multicolumn{2}{c|}{FTD}
		& \multicolumn{2}{c|}{DHD}
		& \multicolumn{2}{c|}{HSTS}
		& \multicolumn{2}{c|}{Fattal}
		& \multicolumn{2}{c}{Average} \\
		\cmidrule{2-11}
		~ 
		& NIMA$\textcolor{cyan!60!blue}{\uparrow}$ & FADE$\textcolor{cyan!60!blue}{\downarrow}$
& NIMA$\textcolor{cyan!60!blue}{\uparrow}$ & FADE$\textcolor{cyan!60!blue}{\downarrow}$
& NIMA$\textcolor{cyan!60!blue}{\uparrow}$ & FADE$\textcolor{cyan!60!blue}{\downarrow}$
& NIMA$\textcolor{cyan!60!blue}{\uparrow}$ & FADE$\textcolor{cyan!60!blue}{\downarrow}$
& NIMA$\textcolor{cyan!60!blue}{\uparrow}$ & FADE$\textcolor{cyan!60!blue}{\downarrow}$ \\
		\midrule
		
		DCP~\cite{he2011single}
		& 5.2225 & \cellcolor{second}\underline{0.3897} & 5.2384 & \cellcolor{second}\underline{0.3563} & 5.1832 & \cellcolor{second}\underline{0.4933} & \cellcolor{second}\underline{5.2750} & \cellcolor{second}\underline{0.2661}
		& 5.2298 & \cellcolor{second}\underline{0.3764} \\

		BCCR~\cite{meng2013efficient}
		& 5.1727 & 0.4203 & 5.1880 & 0.3936 & 5.2131 & 0.5613 & 5.2317 & 0.2964
		& 5.2014 & 0.4179 \\

		CAP~\cite{zhu2015a} 
		& 5.2168 & 0.7442 & 5.2359 & 0.6956 & 5.2408 & 0.8390 & 5.1557 & 0.4714
		& 5.2123 & 0.6876 \\

		MSCNN~\cite{ren2016single} 
		& 5.2735 & 0.7470 & 5.2759 & 0.7264 & 5.2958 & 1.0516 & 5.1945 & 0.4394
		& 5.2600 & 0.7412 \\

		MR~\cite{salazar2018fast} 
		& 5.2617 & 0.4322 & 5.2444 & 0.3902 & 5.2606 & 0.6230 & 5.2601 & 0.2779
		& 5.2567 & 0.4308 \\

		GCP~\cite{ju2020idgcp}
		& 4.8402 & 0.6502 & 4.8685 & 0.6085 & 4.8101 & 0.8169 & 4.7188 & 0.4794
		& 4.8094 & 0.6388 \\

		SBTE~\cite{kim2020fast} 
		& \cellcolor{second}\underline{5.3716} & 0.6226 & \cellcolor{second}\underline{5.3764} & 0.5620 & 5.2872 & 0.8938 & 5.2546 & 0.3730
		& \cellcolor{second}\underline{5.3225} & 0.6129 \\

		CC~\cite{dhara2020color} 
		& 5.2923 & 0.4065 & 5.3008 & 0.3750 & 5.2652 & 0.5672 & 5.2679 & 0.2775
		& 5.2816 & 0.4066 \\

		ROP~\cite{liu2021rank} 
		& 5.3707 & 0.8744 & 5.3740 & 0.7876 & \cellcolor{second}\underline{5.3015} & 1.1223 & 5.1595 & 0.4936
		& 5.3014 & 0.8195 \\

		ROP+~\cite{liu2023rank}
		& 5.0983 & 0.5310 & 5.0957 & 0.4869 & 4.7376 & 0.5837 & 4.8134 & 0.3139
		& 4.9363 & 0.4789 \\

		COA~\cite{ma2025coa} 
		& 5.1287 & 0.7652 & 5.1530 & 0.6472 & 5.2708 & 0.7957 & 5.0704 & 0.3353
		& 5.1557 & 0.6359 \\

		DNMGDT~\cite{su2025real} 
		& 5.0030 & 0.7407 & 4.9731 & 0.6869 & 4.9445 & 1.1915 & 4.8421 & 0.4740
		& 4.9407 & 0.7733 \\

		IDB~\cite{li2025low} 
		& 4.2704 & 1.7605 & 4.1890 & 1.6337 & 5.2027 & 1.9614 & 4.2149 & 0.8568
		& 4.4693 & 1.5531 \\
        
		SDO~\cite{ling2025efficient} 
		& 5.3627 & 0.6533 & 5.3707 & 0.5940 & 5.2664 & 0.8323 & 5.2247 & 0.3843
		& 5.3061 & 0.6150 \\
        
		PDDA~\cite{zhang2026pdda} 
        & 5.3335 & 1.0660 & 5.3591 & 0.9053 & 5.2570 & 1.1870 & 5.2253 & 0.4331 
        & 5.2937 & 0.8979 \\

		SFP
		& \cellcolor{best}\textbf{5.4048} & \cellcolor{best}\textbf{0.3226} 
        & \cellcolor{best}\textbf{5.4240} & \cellcolor{best}\textbf{0.3165} 
        & \cellcolor{best}\textbf{5.3190} & \cellcolor{best}\textbf{0.3934} 
        & \cellcolor{best}\textbf{5.2833} & \cellcolor{best}\textbf{0.2073}
		& \cellcolor{best}\textbf{5.3578} & \cellcolor{best}\textbf{0.3100} \\

		\bottomrule
	\end{tabular}
	\vspace{-0.2cm}
	\label{table:dehazing}
\end{table*}
\begin{figure*}[!ht]
    \centering
    \setlength{\abovecaptionskip}{4pt}  
    \setlength{\belowcaptionskip}{-10pt}  
    \includegraphics[width=7.15in]{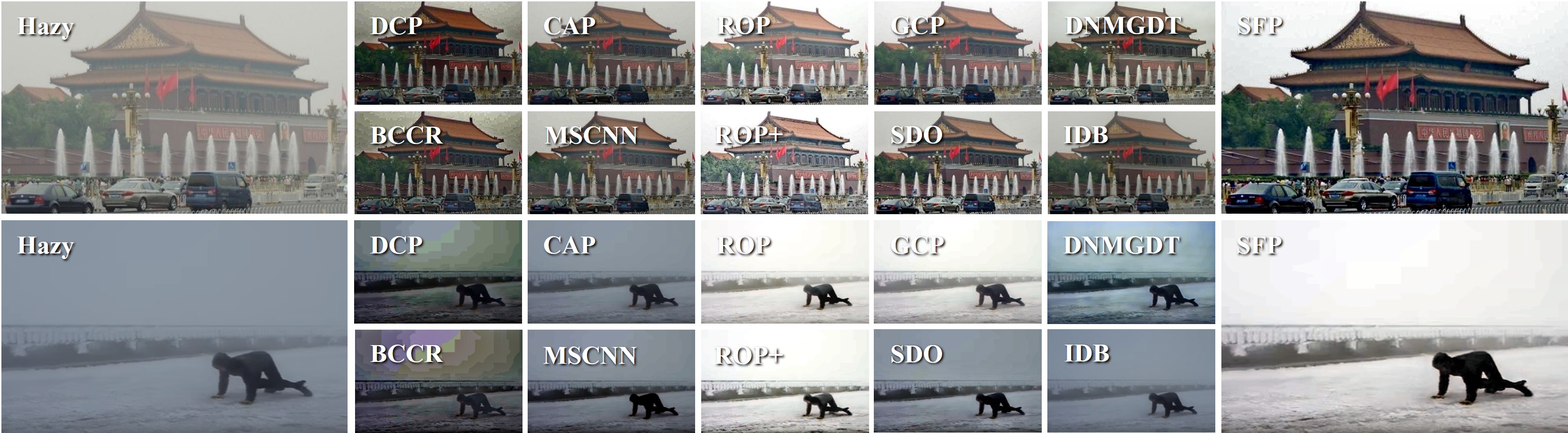}
    \caption{Comparison of state-of-the-art methods on real-world hazy images.   } 
    \label{fig:haze}
\end{figure*}
\begin{figure*}[!t]
    \centering
    \setlength{\abovecaptionskip}{4pt}  
    \setlength{\belowcaptionskip}{-10pt}  
    \includegraphics[width=7.15in]{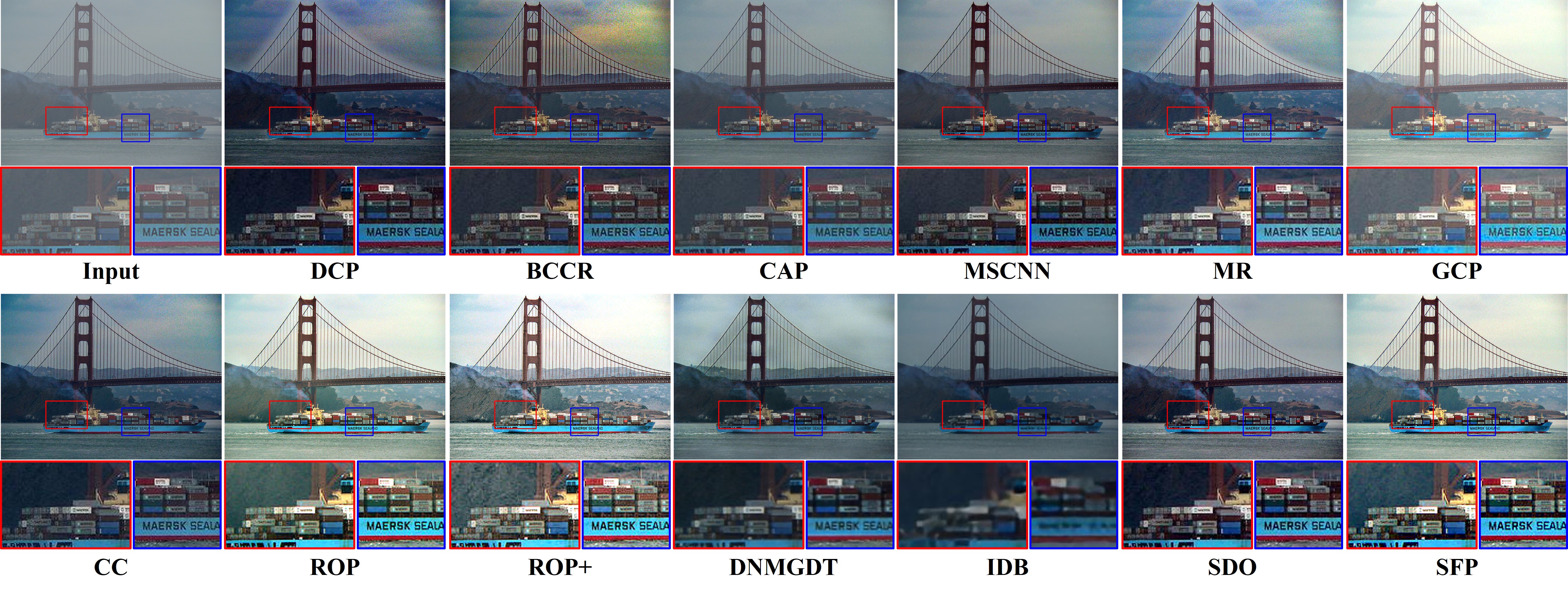}
    \caption{Comparison of detail recovery with state-of-the-art methods on a real-world hazy image.} 
    \label{fig:daytimehaze_detail}
\end{figure*}

\section{Experimental Results and Analysis}
In this section, we first present the experimental settings, followed by objective and visual comparisons with state-of-the-art methods. We then conduct ablation studies to evaluate the contribution of each component. Furthermore, the generalization ability of our algorithm on challenging nighttime hazy images is validated. Finally, we demonstrate the effectiveness of our framework for high-level vision tasks.

\subsection{Experimental Settings}

\textbf{Implementation Details.} All experiments are performed using MATLAB R2021b on a personal computer equipped with an AMD Ryzen 7 6800H processor and 16 GB RAM. Notably, the proposed framework requires no large-scale datasets or extensive computational resources and operates autonomously without adjustable parameters, ensuring adaptability. In the proposed algorithm, the gamma correction and high dynamic range (HDR) compression are applied as post-processing steps to improve visual quality.

\begin{table*}[t]
	\renewcommand\arraystretch{1.2}
	\setlength{\tabcolsep}{3pt}
	\centering
	\caption{Quantitative comparison of the proposed SFP and state-of-the-art methods on two real-world sandstorm image datasets.}
	\vspace{-0.2cm}
	\begin{tabular}{c|c|cccccccccc}
		\toprule
		\multirow{2}{*}{Dataset} & \multirow{2}{*}{Metric}
		& \multicolumn{10}{c}{Methods} \\
		\cmidrule{3-12}
		& 
		& FBE~\cite{fu2014fusion} & TTFIO~\cite{al2016visibility} & GDCP~\cite{peng2018generalization} & ALA~\cite{peng2019image} & HRDCP~\cite{shi2019let} & CVCGCD~\cite{jeon2022sand} & TOE-Net~\cite{gao2023let} & MCR~\cite{al2024increasing}& AOSR-Net~\cite{lu2024aosrnet} & SFP \\
		\midrule
		
		\multirow{2}{*}{DID}
		& NIQE\textcolor{cyan!60!blue}{$\downarrow$}
		& 4.0109 & 4.0512 & 4.1170 & 4.0739 & 4.2869 & 3.9685 & \cellcolor{second}\underline{3.9341} & 4.3374 & 4.1192 & \cellcolor{best}\textbf{3.8680} \\
		
		& FADE\textcolor{cyan!60!blue}{$\downarrow$}
		& 0.9096 & 1.0506 & \cellcolor{second}\underline{0.7968} & 0.8322 & 0.9199 & 1.1570 & 1.0142 & 1.9304 & 1.6890 & \cellcolor{best}\textbf{0.7168} \\
		
		\midrule
		
		\multirow{2}{*}{DAWN}
		& NIQE\textcolor{cyan!60!blue}{$\downarrow$}
		& 5.0274 & 5.3587 & 5.2621 & 5.4837 & \cellcolor{second}\underline{4.7314} & 5.1482 & 5.0295 & 5.8686 & 5.3988 & \cellcolor{best}\textbf{4.4371} \\
		
		& FADE\textcolor{cyan!60!blue}{$\downarrow$}
		& 1.6171 & 1.5795 & \cellcolor{second}\underline{1.1756} & 1.2328 & 1.7124 & 2.0817 & 1.4839 & 2.8632 & 2.5490 & \cellcolor{best}\textbf{1.0398} \\
		
		\midrule
		
		\multirow{2}{*}{Average}
		& NIQE\textcolor{cyan!60!blue}{$\downarrow$}
		& 4.5192 & 4.7050 & 4.6896 & 4.7788 & 4.5092 & 4.5584 & \cellcolor{second}\underline{4.4818} & 5.1030 & 4.7590 & \cellcolor{best}\textbf{4.1526} \\
		
		& FADE\textcolor{cyan!60!blue}{$\downarrow$}
		& 1.2634 & 1.3151 & \cellcolor{second}\underline{0.9862} & 1.0325 & 1.3162 & 1.6194 & 1.2491 & 2.3968 & 2.1190 & \cellcolor{best}\textbf{0.8783} \\
		
		\bottomrule
	\end{tabular}
	\vspace{-0.2cm}
	\label{table:sandstorm}
\end{table*}

\begin{figure*}[!ht]
    \centering
    \setlength{\abovecaptionskip}{4pt}  
    \setlength{\belowcaptionskip}{-10pt}  
    \includegraphics[width=7.15in]{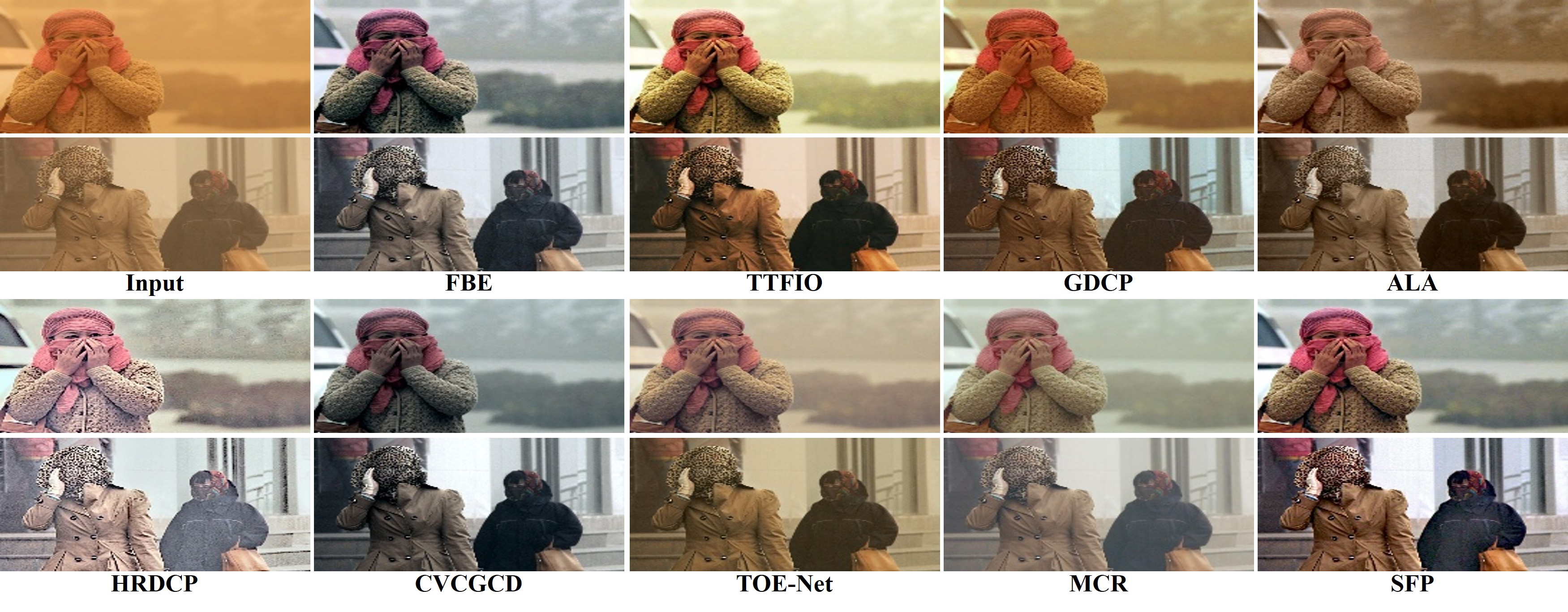}
    \caption{Comparison of state-of-the-art methods on real-world sandstorm images. } 
    \label{fig:sandstorm}
\end{figure*}
\begin{figure*}[!ht]
    \centering
    \setlength{\abovecaptionskip}{4pt}  
    \setlength{\belowcaptionskip}{-10pt}  
    \includegraphics[width=7.15in]{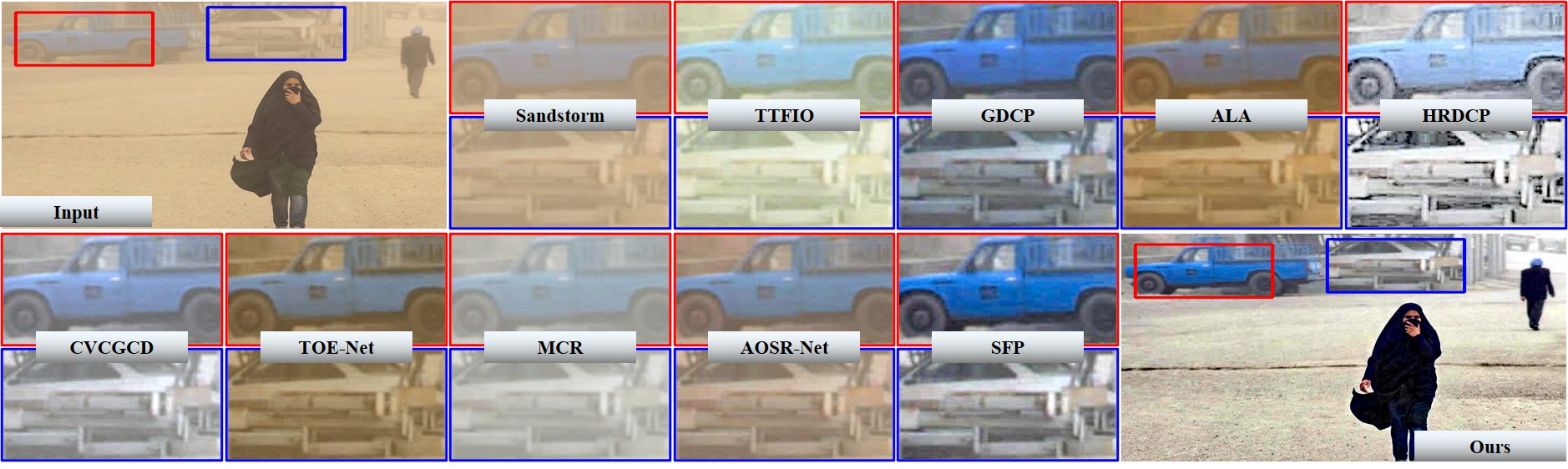}
    \caption{Comparison of detail recovery with state-of-the-art methods on a real-world sandstorm image.} 
    \label{fig:sandstorm_detail}
\end{figure*}

\begin{table*}[t]
	\renewcommand\arraystretch{1.2}
	\setlength{\tabcolsep}{9pt}
	\centering
	\caption{Quantitative comparison of the proposed SFP and state-of-the-art methods on four real-world underwater image datasets.}
	\vspace{-0.2cm}
	\begin{tabular}{c|cc|cc|cc|cc|cc}
		\toprule
		\multirow{2}{*}{Methods}
		& \multicolumn{2}{c|}{UIEB}
		& \multicolumn{2}{c|}{U45}
		& \multicolumn{2}{c|}{UCCS}
		& \multicolumn{2}{c|}{UHTS}
		& \multicolumn{2}{c}{Average} \\
		\cmidrule{2-11}
		~
		& NIMA$\textcolor{cyan!60!blue}{\uparrow}$ & UCIEQ$\textcolor{cyan!60!blue}{\uparrow}$
& NIMA$\textcolor{cyan!60!blue}{\uparrow}$ & UCIEQ$\textcolor{cyan!60!blue}{\uparrow}$
& NIMA$\textcolor{cyan!60!blue}{\uparrow}$ & UCIEQ$\textcolor{cyan!60!blue}{\uparrow}$
& NIMA$\textcolor{cyan!60!blue}{\uparrow}$ & UCIEQ$\textcolor{cyan!60!blue}{\uparrow}$
& NIMA$\textcolor{cyan!60!blue}{\uparrow}$ & UCIEQ$\textcolor{cyan!60!blue}{\uparrow}$ \\
		\midrule

		IBLA~\cite{peng2017underwater}
		& 4.5914 & 0.5980 & 4.0642 & 0.5656 & 3.5664 & 0.5003 & \cellcolor{second}\underline{3.6599} & 0.5413
		& 3.9705 & 0.5513 \\

		GDCP~\cite{peng2018generalization}
		& 4.4398 & 0.5959 & 4.0191 & 0.5774 & 3.4312 & 0.5385 & 3.5453 & 0.5387
		& 3.8589 & 0.5626 \\

		CBAF~\cite{ancuti2018color}
		& 4.5436 & 0.5510 & 4.0154 & 0.5291 & 3.4154 & 0.4762 & 3.5146 & 0.5045
		& 3.8723 & 0.5152 \\

		ROP~\cite{liu2021rank}
		& 4.6067 & 0.6302 & 4.0551 & 0.6193 & \cellcolor{second}\underline{3.6233} & \cellcolor{second}\underline{0.6243} & 3.6015 & \cellcolor{second}\underline{0.6139}
		& 3.9717 & 0.6219 \\

		TEBCF~\cite{yuan2021tebcf}
		& 4.6472 & 0.6296 & 4.0854 & 0.6237 & 3.3918 & 0.6071 & 3.4150 & 0.5945
		& 3.8849 & 0.6137 \\

		ROP+~\cite{liu2023rank}
		& 4.2856 & 0.6391 & 3.9358 & 0.6355 & 3.4505 & 0.6194 & 3.4803 & \cellcolor{second}\underline{0.6139}
		& 3.7881 & \cellcolor{second}\underline{0.6270} \\

		WWPF~\cite{zhang2024wwpf}
		& 4.4912 & 0.6127 & 3.8735 & 0.5986 & 3.3424 & 0.5853 & 3.4190 & 0.5874
		& 3.7815 & 0.5960 \\

		PCFB~\cite{zhang2024pcfb}
		& 4.7158 & 0.6379 & 4.2181 & 0.6347 & 3.4064 & 0.5730 & 3.5838 & 0.5811
		& 3.9810 & 0.6067 \\

		C3HLM~\cite{wang2024underwater}
		& 4.6952 & 0.6079 & 4.0713 & 0.6035 & 3.4162 & 0.6045 & 3.4512 & 0.5959
		& 3.9085 & 0.6030 \\

		AFWF~\cite{zhang2025underwater}
		& 4.2222 & 0.5323 & 3.7209 & 0.5140 & 3.1396 & 0.4908 & 3.2826 & 0.5082
		& 3.5913 & 0.5113 \\

		WFAC~\cite{zhang2025wfac}
		& \cellcolor{second}\underline{4.7379} & 0.6096 & \cellcolor{best}\textbf{4.2897} & 0.6053 & 3.4832 & 0.6025 & 3.5379 & 0.5992
		& \cellcolor{second}\underline{4.0122} & 0.6042 \\

		UDHTV~\cite{li2025dual}
		& 4.6086 & \cellcolor{second}\underline{0.6474} & 4.0118 & 0.6293 & 3.4551 & 0.6123 & 3.4630 & 0.6084
		& 3.8846 & 0.6244 \\

		ALSP~\cite{he2025alsp}
		& 4.6372 & 0.6359 & 4.1565 & \cellcolor{second}\underline{0.6364} & 3.3911 & 0.6132 & 3.5244 & 0.6089
		& 3.9273 & 0.6236 \\

		SEA-PACE~\cite{wang2026sea}
		& 3.7840 & 0.6150 & 4.0883 & 0.6126 & 3.4161 & 0.5419 & 3.4382 & 0.5644
		& 3.6817 & 0.5835 \\

		DTI-UIE~\cite{lin2026downstream}
		& 3.7803 & 0.5734 & 3.9697 & 0.5628 & 3.3392 & 0.5185 & 3.3559 & 0.5376
		& 3.6113 & 0.5481 \\

		SFP
		& \cellcolor{best}\textbf{4.7399} & \cellcolor{best}\textbf{0.6864} & \cellcolor{second}\underline{4.2541} & \cellcolor{best}\textbf{0.6887} & \cellcolor{best}\textbf{3.6322} & \cellcolor{best}\textbf{0.6470} & \cellcolor{best}\textbf{3.7260} & \cellcolor{best}\textbf{0.6538}
		& \cellcolor{best}\textbf{4.0881} & \cellcolor{best}\textbf{0.6690} \\
		\bottomrule
	\end{tabular}
	\vspace{-0.2cm}
	\label{table:underwater}
\end{table*}
\begin{figure*}[!ht]
    \centering
    \setlength{\abovecaptionskip}{4pt}  
    \setlength{\belowcaptionskip}{-10pt}  
    \includegraphics[width=7.15in]{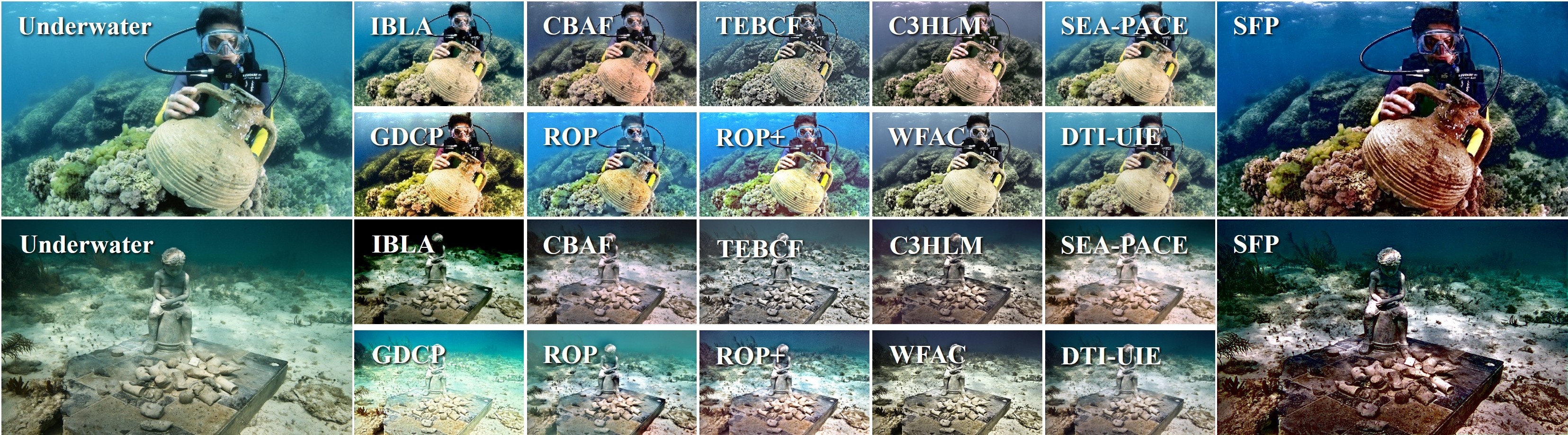}
    \caption{Comparison of state-of-the-art methods on real-world underwater degraded images.} 
    \label{fig:underwater}
\end{figure*}
\begin{figure*}[!ht]
    \centering
    \setlength{\abovecaptionskip}{4pt}  
    \setlength{\belowcaptionskip}{-10pt}  
    \includegraphics[width=7.15in]{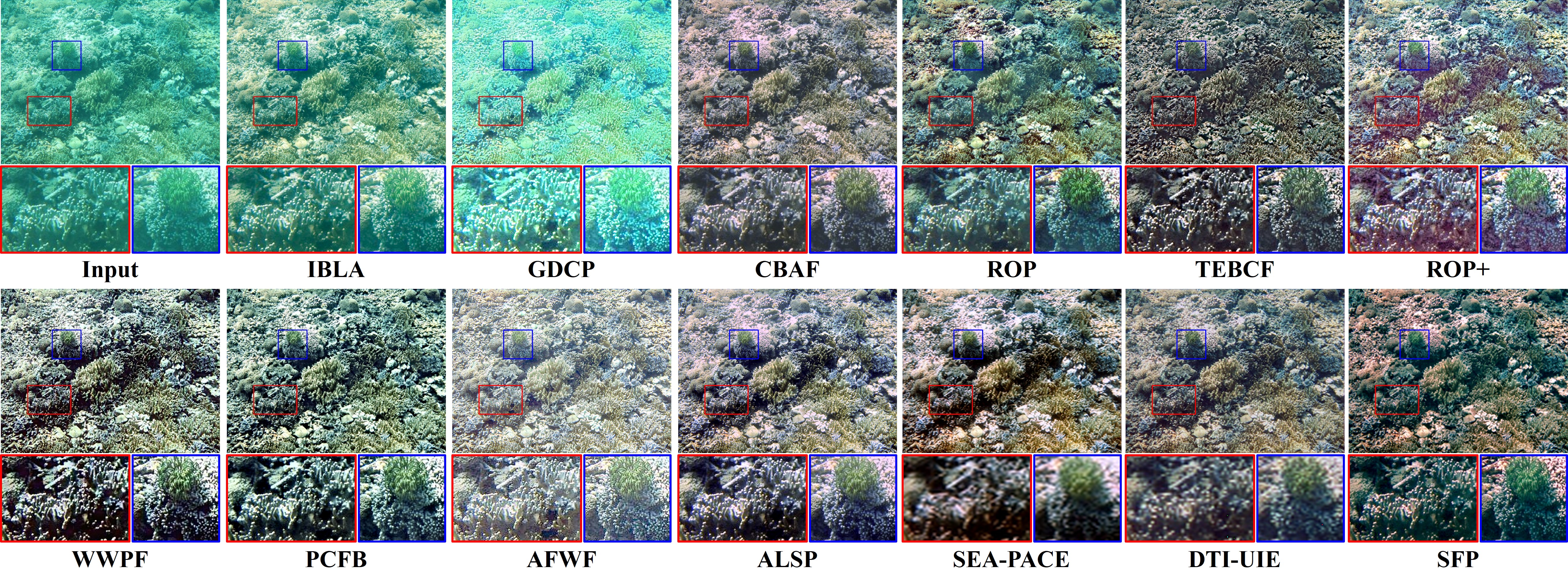}
    \caption{Comparison of detail recovery with state-of-the-art methods on a real-world underwater image.} 
    \label{fig:underwater_detail}
\end{figure*}
\textbf{Real-World Datasets.} To comprehensively evaluate the proposed SFP, we conduct experiments on a diverse collection of real-world degraded image datasets, including four hazy image datasets (FTD~\cite{choi2015referenceless}, DHQ~\cite{min2019objective}, HSTS~\cite{li2019benchmarking}, and Fattal~\cite{fattal2014dehazing}), two sandstorm image datasets (DID~\cite{bartani2022adaptive} and DAWN~\cite{hassaballah2020vehicle}), four underwater image datasets (UIEB~\cite{li2019underwater}, U45~\cite{li2019fusion}, UCCS~\cite{liu2020real}, and UHTS~\cite{liu2020real}), and three remote sensing hazy image datasets (UAV~\cite{zheng2023uav}, RRSHID~\cite{he2023remote}, and RRSD~\cite{wen2023encoder}).
Specifically, FTD contains 500 outdoor images with varying haze densities, DHQ provides 250 images covering diverse scenes, HSTS includes 10 real-world samples, and Fattal offers 31 images for evaluation. 
For sandstorm scenarios, DID consists of 140 images collected from different regions worldwide, and DAWN contains 323 samples captured in real traffic environments. 
For underwater image enhancement, UIEB includes 950 images with varying degradation levels, U45 is a small-scale test set with 45 images, UCCS contains 300 images for color correction evaluation, and UHTS provides 300 images for downstream task assessment. 
For remote sensing dehazing, UAV is a partially public dataset containing 50 images captured by unmanned aerial vehicles, RRSHID contains 277 samples affected by haze, and RRSD includes 300 remote sensing hazy images.

In order to further evaluate the robustness of the proposed SFP under mixed degradation conditions, we  adopt the RUSH dataset~\cite{liu2023rank}, which is specifically designed to contain heterogeneous real-world degradations. Unlike single-type degradation datasets, RUSH contains a mixture of haze, sandstorm, and underwater images, providing a more challenging benchmark for evaluating the evaluating the robustness of different methods under mixed degradation conditions. Specifically, it includes 100 hazy images, 200 sandstorm images, and 50 underwater images. 

Finally, to evaluate the generalization capability of the proposed SFP under more complex real-world conditions, two real-world nighttime hazy image datasets, namely NHRW~\cite{zhang2020nighttime} and RealNightHaze~\cite{jin2023enhancing}, are employed for evaluation. NHRW contains 150 images with varying degradation levels, while RealNightHaze provides 443 real-world nighttime hazy images collected from the Internet.


\textbf{Compared Methods.}  To conduct a comprehensive and objective evaluation, we selected more than $50$ classic and recent state-of-the-art methods for both quantitative and qualitative comparisons. Specifically, these methods include 12 dehazing methods, 7 sandstorm enhancement methods, 11 underwater enhancement methods, 8 remote sensing haze removal methods, 6 nighttime dehazing methods, and 9 multi-scenario restoration methods. All compared methods are evaluated using their officially released implementations and pretrained weights, except for DCP~\cite{he2011single} and CBAF~\cite{ancuti2018color}, which are implemented based on publicly available repositories\footnote{\url{https://github.com/sjtrny/Dark-Channel-Haze-Removal}}$^{,}$\footnote{\url{https://github.com/fergaletto/Color-Balance-and-fusion-for-underwater-image-enhancement.-.}}.

\begin{table*}[t]
	\renewcommand\arraystretch{1.2}
	\setlength{\tabcolsep}{4pt}
	\centering
	\caption{Quantitative comparison of the proposed SFP and state-of-the-art methods on three real-world remote sensing image datasets.}
	\vspace{-0.2cm}
	\begin{tabular}{c|c|cccccccccc}
		\toprule
		\multirow{2}{*}{Dataset} & \multirow{2}{*}{Metric}
		& \multicolumn{10}{c}{Methods} \\
		\cmidrule{3-12}
		&
		& SDCP~\cite{li2018haze} & HTM~\cite{liu2017haze} & HALP~\cite{he2023remote} & IDE~\cite{ju2021ide}& EMPF-Net~\cite{wen2023encoder} & SRD~\cite{he2023srd} & ROP~\cite{liu2021rank} & SLP~\cite{ling2023single} & IHDCP~\cite{liu2026ihdcp} & SFP \\
		\midrule
		
		\multirow{2}{*}{UAV}
		& NIMA\textcolor{cyan!60!blue}{$\uparrow$}
		& 3.4507 & 3.3262 & 3.4201 & 3.5673 & \cellcolor{second}\underline{3.6098} & 3.4022 & 3.4260 & 3.5467 & 3.3679 & \cellcolor{best}\textbf{3.6149} \\
		
		& FADE\textcolor{cyan!60!blue}{$\downarrow$}
		& 0.4657 & 3.8765 & 1.1276 & \cellcolor{second}\underline{0.4531} & 2.8620 & 1.6688 & 0.5587 & 0.4708 & 1.2682 & \cellcolor{best}\textbf{0.4397} \\
		
		\midrule
		
		\multirow{2}{*}{RRSD}
		& NIMA\textcolor{cyan!60!blue}{$\uparrow$}
		& 4.6121 & 4.2807 & 4.4968 & \cellcolor{second}\underline{4.6325} & 4.4540 & 4.5626 & 4.6106 & 4.6142 & 4.5638 & \cellcolor{best}\textbf{4.6916} \\
		
		& FADE\textcolor{cyan!60!blue}{$\downarrow$}
		& \cellcolor{second}\underline{0.2918} & 1.2583 & 0.5284 & 0.3574 & 0.7669 & 0.6036 & 0.5083 & 0.4024 & 0.6299 & \cellcolor{best}\textbf{0.2688} \\
		
		\midrule
		
		\multirow{2}{*}{RRSHID}
		& NIMA\textcolor{cyan!60!blue}{$\uparrow$}
		& \cellcolor{second}\underline{4.6883} & 4.2449 & 4.5691 & 4.6482 & 4.4787 & 4.6339 & 4.6531 & 4.5858 & 4.5768 & \cellcolor{best}\textbf{4.7146} \\
		
		& FADE\textcolor{cyan!60!blue}{$\downarrow$}
		& \cellcolor{second}\underline{0.2300} & 0.9271 & 0.3728 & 0.3010 & 0.6450 & 0.4013 & 0.3728 & 0.3039 & 0.4032 & \cellcolor{best}\textbf{0.1763} \\
		
		\midrule
		
		\multirow{2}{*}{Average}
		& NIMA\textcolor{cyan!60!blue}{$\uparrow$}
		& 4.2504 & 3.9506 & 4.1620 & \cellcolor{second}\underline{4.2827} & 4.1808 & 4.1996 & 4.2299 & 4.2489 & 4.1695 & \cellcolor{best}\textbf{4.3404} \\
		
		& FADE\textcolor{cyan!60!blue}{$\downarrow$}
		& \cellcolor{second}\underline{0.3292} & 2.0206 & 0.6763 & 0.3705 & 1.4246 & 0.8912 & 0.4799 & 0.3924 & 0.7671 & \cellcolor{best}\textbf{0.2949} \\
		
		\bottomrule
	\end{tabular}
	\vspace{-0.2cm}
	\label{table:uav_rrs}
\end{table*}
\begin{figure*}[!t]
    \centering
    \setlength{\abovecaptionskip}{4pt}  
    \setlength{\belowcaptionskip}{-10pt}  
    \includegraphics[width=7.15in]{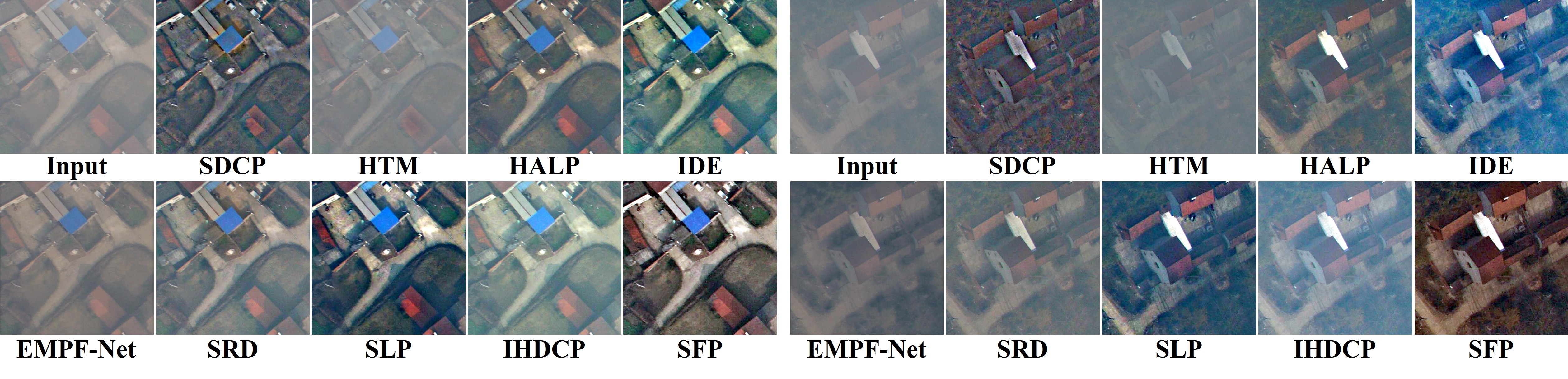}
    \caption{Comparison of state-of-the-art methods on real-world remote sensing degraded images.} 
    \label{fig:remote_sensing}
\end{figure*}
\begin{figure*}[!t]
    \centering
    \setlength{\abovecaptionskip}{4pt}  
    \setlength{\belowcaptionskip}{-10pt}  
    \includegraphics[width=7.15in]{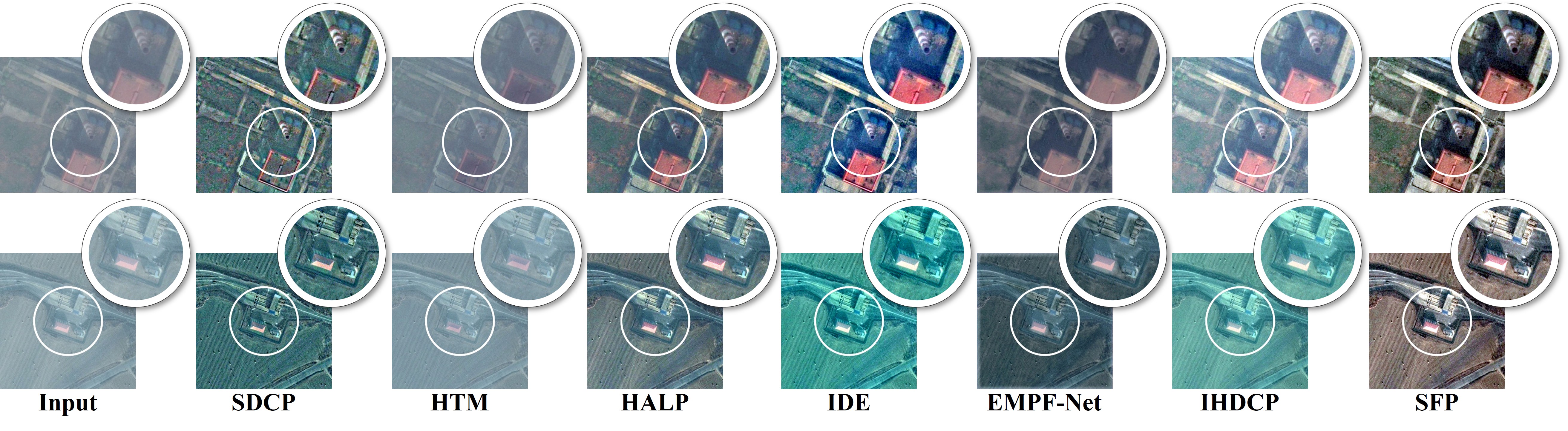}
    \caption{Comparison of detail recovery with state-of-the-art methods on a real-world remote sensing degraded image.} 
    \label{fig:remote_sensing_detail}
\end{figure*}

\textbf{Evaluation Metrics. }To avoid subjective bias, we employ four no-reference metrics, namely NIMA~\cite{talebi2018nima}, NIQE~\cite{mittal2013making}, FADE~\cite{choi2015referenceless}, and UCIQE~\cite{yang2015underwater}, to quantitatively evaluate the proposed SFP and other state-of-the-art methods. Specifically, NIMA is adopted to assess the aesthetic quality of images, NIQE evaluates image naturalness, FADE measures the haze density of a scene, and UCIQE provides a comprehensive assessment of color balance and visual clarity for underwater images. FADE is evaluated using its official implementation, while the implementations of NIMA and NIQE are obtained from a publicly available open-source library\footnote{\url{https://github.com/chaofengc/IQA-PyTorch}}, and UCIQE is implemented based on a public repository\footnote{\url{https://github.com/JOU-UIP/UCIQE}}. Notably, higher values of NIMA and UCIQE indicate better image quality, whereas lower values of FADE and NIQE are preferred.

\begin{table*}[t]
\renewcommand\arraystretch{1.2}
\setlength{\tabcolsep}{1pt}
\centering
\caption{Quantitative comparison of the proposed SFP and state-of-the-art methods on the RUSH dataset.}
\vspace{-0.2cm}
 \resizebox{7.15in}{!}{
\begin{tabular}{c|c|cccccccccccc}
\toprule

Dataset & Methods
& GDCP~\cite{peng2018generalization} & ROP~\cite{liu2021rank} & ROP+~\cite{liu2023rank} & SDO~\cite{ling2025efficient} & DNMGDT~\cite{su2025real} & IDB~\cite{li2025low} & GCP~\cite{ju2019idgcp} & GLP~\cite{he2026efficient} & CAP~\cite{zhu2015a} & MSCNN~\cite{ren2016single} & SBTE~\cite{kim2020fast} & SFP \\ \midrule

\multirow{2}{*}{Hazy} & NIMA\textcolor{cyan!60!blue}{$\uparrow$}
& 4.8866 & 4.8710 & 4.7632 & 4.9793 & 4.7744 & 4.3530 & 4.6045 & \cellcolor{second}\underline{4.9967} & 4.7034 & 4.8119 & 4.9461 & \cellcolor{best}\textbf{5.0503} \\

& FADE\textcolor{cyan!60!blue}{$\downarrow$}
& 1.1187 & 1.3970 & \cellcolor{second}\underline{0.9537} & 1.0549 & 1.1386 & 2.4083 & 2.0748 & 1.0938 & 1.7883 & 1.5192 & 1.0775 & \cellcolor{best}\textbf{0.9210} \\

\midrule

Dataset & Methods
& GDCP~\cite{peng2018generalization} & ROP~\cite{liu2021rank} & FBE~\cite{fu2014fusion} & TTFIO~\cite{al2016visibility} & AOSR-Net~\cite{lu2024aosrnet} & TOE-Net~\cite{gao2023let} & ALA~\cite{peng2019image} & MCR~\cite{al2024increasing} & CVCGCD~\cite{jeon2022sand} & OATF~\cite{bartani2022adaptive} & HRDCP~\cite{shi2019let} & SFP \\ \midrule

\multirow{2}{*}{Sandstorm} & NIQE\textcolor{cyan!60!blue}{$\downarrow$}
& 3.9048 & 3.6963 & 3.7286 & 3.8955 & 4.0894 & 3.9245 & 4.0276 & 4.2470 & 3.9179 & \cellcolor{best}\textbf{3.5788} & 3.9243 & \cellcolor{second}\underline{3.6606} \\

& FADE\textcolor{cyan!60!blue}{$\downarrow$}
& \cellcolor{best}\textbf{1.0089} & 1.4611 & 1.3255 & 1.5422 & 1.9544 & 1.5422 & 1.1533 & 2.1938 & 1.5622 & 1.1351 & 1.3336 & \cellcolor{second}\underline{1.0623} \\

\midrule

Dataset & Methods
& GDCP~\cite{peng2018generalization} & ROP~\cite{liu2021rank} & ROP+~\cite{liu2023rank} & ALSP~\cite{he2025alsp} & PCFB~\cite{zhang2024pcfb} & C3HLM~\cite{wang2024underwater} & AFWF~\cite{zhang2025underwater} & WWPF~\cite{zhang2024wwpf} & CBAF~\cite{ancuti2018color} & SEA-PACE~\cite{wang2026sea} & DTI-UIE~\cite{lin2026downstream} & SFP \\ \midrule

\multirow{2}{*}{Underwater} & NIMA\textcolor{cyan!60!blue}{$\uparrow$}
& 4.5846 & 4.8160 & 4.4766 & 4.8468 & 4.9192 & \cellcolor{second}\underline{4.9434} & 4.5232 & 4.7644 & 4.7490 & 3.8618 & 3.8561 & \cellcolor{best}\textbf{4.9568} \\

& UCIEQ\textcolor{cyan!60!blue}{$\uparrow$}
& 0.5661 & 0.6392 & \cellcolor{second}\underline{0.6446} & 0.6376 & 0.6356 & 0.6099 & 0.5249 & 0.6112 & 0.5338 & 0.6159 & 0.5646 & \cellcolor{best}\textbf{0.6827} \\

\bottomrule
\end{tabular}
}
\vspace{-0.2cm}
\label{tab:rush}
\end{table*}

\subsection{Objective Comparisons with State-of-the-Art Methods}
To quantitatively evaluate the effectiveness of the proposed SFP across diverse real-world degradation scenarios, we conduct extensive comparisons with more than $50$ state-of-the-art (SOTA) methods on multiple benchmark datasets, including haze, sandstorm, underwater, and remote sensing scene recovery tasks, as summarized in Tables~\ref{table:dehazing}-\ref{tab:rush}.

\subsubsection{Evaluation on Real-World Hazy Datasets} As shown in Table~\ref{table:dehazing}, the proposed SFP achieves the best performance across all four real-world daytime hazy datasets, namely FTD, DHQ, HSTS, and Fattal. Specifically, our SFP improves the average NIMA score by approximately $0.66\%$ and reduces the average FADE value by approximately $17.64\%$. On the HSTS dataset, the FADE metric is further reduced by approximately $20.25\%$. 
\subsubsection{Evaluation on Real-World Sandstorm Datasets}
Table~\ref{table:sandstorm} presents the quantitative comparison of the proposed SFP on two real-world sandstorm datasets. According to Table~\ref{table:sandstorm}, our SFP improves the average NIQE and FADE scores by approximately $7.35\%$ and $10.94\%$, respectively, while achieving a reduction of approximately $6.22\%$ in NIQE metric on the DAWN dataset.
\subsubsection{Evaluation on Real-World Underwater Datasets}
Table~\ref{table:underwater} shows the quantitative comparison between the proposed SFP and several SOTA underwater image enhancement methods on four real-world underwater datasets. The results indicate that the proposed SFP achieves the best performance in most cases. Specifically, the proposed framework improves the average NIMA and UCIQE scores by approximately $1.89\%$ and $6.70\%$, respectively. On the UIEB dataset, the UCIQE score is further improved by approximately $6.02\%$.
\subsubsection{Evaluation on Real-World Remote Sensing Hazy Datasets}
Table~\ref{table:uav_rrs} reports the quantitative results of the proposed SFP and competing methods on real-world remote sensing hazy datasets. In particular, our SFP decreases the average FADE value by approximately $10.42\%$. A more pronounced improvement can be observed on the RRSHID dataset, where the FADE metric is reduced by approximately $23.35\%$, indicating the effectiveness of the proposed framework for remote sensing haze removal.
\subsubsection{Evaluation on RUSH Dataset}
To demonstrate the capability of the proposed method in simultaneously handling various scattering-induced degradations, we further conduct quantitative comparisons on the real-world mixed degradation benchmark RUSH~\cite{liu2023rank}, as summarized in Table~\ref{tab:rush}. The results show that the proposed SFP achieves competitive performance across different degradation scenarios, including hazy, sandstorm, and underwater environments. Specifically, on the hazy subset, SFP obtains the best NIMA and FADE scores, improving the NIMA metric by approximately $1.07\%$ and reducing the FADE value by approximately $3.43\%$ compared with the second-best results. On the underwater subset, SFP also achieves the best overall performance, with improvements of approximately $0.27\%$ in NIMA and $5.91\%$ in UCIQE.

 In summary, these consistent improvements across diverse datasets and evaluation metrics demonstrate that the proposed SFP effectively enhances image quality and exhibits favorable robustness under various real-world degradation scenarios.

\subsection{Visual Comparisons}
Fig.~\ref{fig:haze} shows the overall dehazing performance and Fig.~\ref{fig:daytimehaze_detail} further highlights local detail restoration capability. 
As depicted in Fig.~\ref{fig:haze}, DCP and BCCR tend to over-enhance the scenes, leading to noticeable halo artifacts in smooth regions such as the sky. CAP and MSCNN exhibit insufficient dehazing capability, where residual haze remains and the overall scene still appears unclear. ROP, ROP+, and GCP often produce overly bright results, resulting in unnatural illumination and degraded visual realism. DNMGDT fails to effectively correct the color distortion caused by scattering effects, leading to noticeable color shifts in the restored results. IDB suffers from blurred structural details, as further evidenced in Fig.~\ref{fig:daytimehaze_detail}, and SDO shows limited ability in restoring fine details in dark regions. In contrast, the proposed method effectively removes haze while maintaining well-balanced brightness and natural color appearance. More importantly, it preserves fine structures in both bright and dark regions, particularly demonstrating good capability in recovering details in low-illumination areas.

For severely color-distorted sandstorm images, Fig.~\ref{fig:sandstorm} presents the overall recovered results, and Fig.~\ref{fig:sandstorm_detail} provides zoomed-in views for local detail comparison.
As shown in Fig.~\ref{fig:sandstorm}, TTFIO, GDCP, ALA, TOE-Net, and MCR exhibit limited restoration capability on real-world sandstorm images, where noticeable color shifts still remain in the recovered results. Although FBE, HRDCP, and CVCGCD can effectively suppress the yellowish sandstorm coloration and perform color correction to some extent, the recovered results tend to suffer from low color saturation, making the overall scenes appear visually dull and less natural.
In contrast, the proposed SFP is able to effectively remove sandstorm-induced color bias while faithfully restoring the intrinsic scene colors. Moreover, as further evidenced in Fig.~\ref{fig:sandstorm_detail}, the proposed algorithm produces more realistic local structures with clearer textures and better preserved scene details.

Underwater images are often severely degraded due to light scattering and absorption in water, resulting in not only significant color distortion but also blurred structural details. As shown in Fig.~\ref{fig:underwater}, although most existing underwater image enhancement methods can effectively improve contrast to some extent, they generally fail to properly correct severe color degradation and produce clear details in the restored results.
Moreover, as further illustrated in the zoomed-in comparisons in Fig.~\ref{fig:underwater_detail}, especially in heavily degraded regions, the proposed SFP significantly enhances image contrast while simultaneously producing more realistic and faithful color restoration. In particular, the enlarged regions demonstrate clearer structures and more natural appearance compared to competing methods.
Thus, these results demonstrate that the proposed SFP is also well-suited for underwater image quality enhancement, effectively improving contrast, color fidelity, and structural detail preservation under underwater degradation conditions.

\begin{table*}[!t]
    \centering
	\renewcommand\arraystretch{1.2}
	\setlength{\tabcolsep}{5pt}
    \caption{Ablation study of individual modules on four real-world datasets}
    
    \begin{tabular}{c|cccc|cc|cc|cc|cc|cc}
    \toprule
    \multirow{2}{*}{Component} 
    & \multirow{2}{*}{SDP} 
    & \multirow{2}{*}{FDP} 
    & \multirow{2}{*}{Fusion} 
    & \multirow{2}{*}{PP} 
    & \multicolumn{2}{c|}{Fattal} 
    & \multicolumn{2}{c|}{FTD} 
    & \multicolumn{2}{c|}{HSTS} 
    & \multicolumn{2}{c|}{DHQ}
    & \multicolumn{2}{c}{Average} \\

    & & & & 
    & NIMA\textcolor{cyan!60!blue}{$\uparrow$} & FADE\textcolor{cyan!60!blue}{$\downarrow$}
    & NIMA\textcolor{cyan!60!blue}{$\uparrow$} & FADE\textcolor{cyan!60!blue}{$\downarrow$}
    & NIMA\textcolor{cyan!60!blue}{$\uparrow$} & FADE\textcolor{cyan!60!blue}{$\downarrow$}
    & NIMA\textcolor{cyan!60!blue}{$\uparrow$} & FADE\textcolor{cyan!60!blue}{$\downarrow$}
    & NIMA\textcolor{cyan!60!blue}{$\uparrow$} & FADE\textcolor{cyan!60!blue}{$\downarrow$} \\
    \midrule

    w/o SDP 
    & \faTimes & \faCheck & \faCheck & \faCheck 
    & 5.1023 & 0.4611 
    & 5.3022 & 0.7985 
    & 5.0657 & 0.9978 
    & 5.3076 & 0.7427
    & 5.1945 & 0.7500 \\ 

    w/o FDP 
    & \faCheck & \faTimes & \faCheck & \faCheck 
    & 5.0546 & 0.4274 
    & 5.2643 & 0.6213 
    & 4.9475 & 0.8275 
    & 5.2580 & 0.5725
    & 5.1311 & 0.6122 \\ 

    w/o Fusion 
    & \faCheck & \faCheck & \faTimes & \faCheck 
    & 5.1670 & 0.4101 
    & 5.3779 & 0.6937 
    & 5.2319 & 0.9318 
    & 5.3816 & 0.6436
    & 5.2896 & 0.6698 \\ 

    w/o PP 
    & \faCheck & \faCheck & \faCheck & \faTimes 
    & \underline{\cellcolor{second}5.2078} & \underline{\cellcolor{second}0.2155} 
    & \underline{\cellcolor{second}5.3486} & \underline{\cellcolor{second}0.3245} 
    & \underline{\cellcolor{second}5.2519} & \underline{\cellcolor{second}0.3726} 
    & \underline{\cellcolor{second}5.3636} & \underline{\cellcolor{second}0.3283}
    & \underline{\cellcolor{second}5.2930} & \underline{\cellcolor{second}0.3102} \\ 

    \midrule

    Ours 
    & \faCheck & \faCheck & \faCheck & \faCheck 
    & \textbf{\cellcolor{best}5.2833} & \textbf{\cellcolor{best}0.2073} 
    & \textbf{\cellcolor{best}5.4048} & \textbf{\cellcolor{best}0.3226} 
    & \textbf{\cellcolor{best}5.3190} & \textbf{\cellcolor{best}0.3934} 
    & \textbf{\cellcolor{best}5.4240} & \textbf{\cellcolor{best}0.3165}
    & \textbf{\cellcolor{best}5.3578} & \textbf{\cellcolor{best}0.3100} \\ 

    \bottomrule
    \end{tabular}%
    \vspace{-0.2cm}
    \label{tab:ablation}
\end{table*}

\begin{figure}[!t]
 \setlength{\abovecaptionskip}{2pt}   
    \setlength{\belowcaptionskip}{-13pt}  
    \centering    
    \includegraphics[width=3.5in]{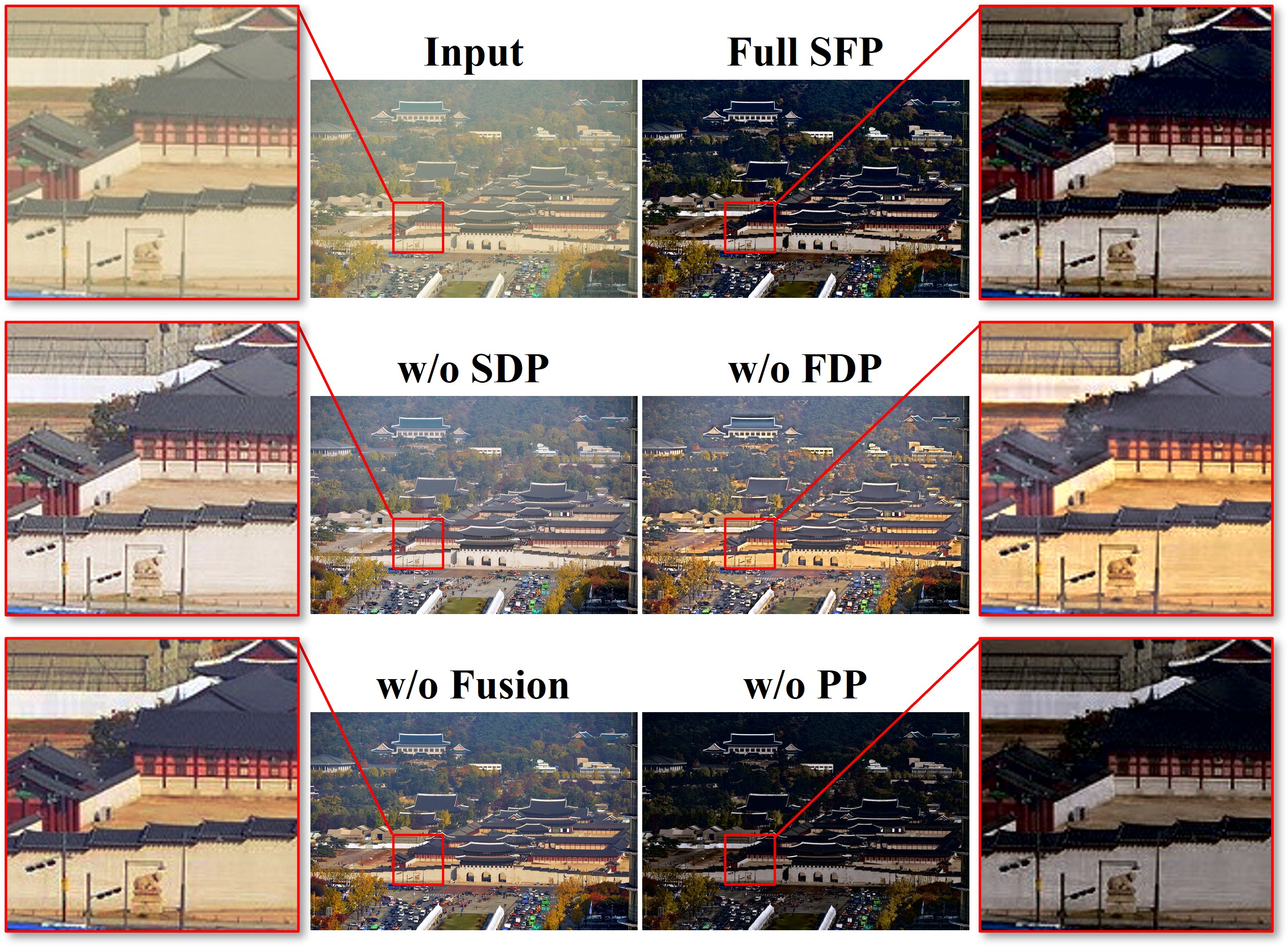}
    \caption{Ablation study of each component in the proposed framework.}
    \label{fig:ablation}
\end{figure}

For remote sensing hazy scenes, the scattering effects caused by clouds and haze not only degrade image clarity but also introduce noticeable color distortions. As shown in Fig.~\ref{fig:remote_sensing}, most existing methods, such as SDCP, IDE, and SLP, can effectively enhance image contrast to some extent. However, as further illustrated in Fig.~\ref{fig:remote_sensing_detail}, the restored results often suffer from unnatural color reproduction and limited visual fidelity.
By comparison, existing methods primarily focus on contrast enhancement, the proposed SFP explicitly models degradation in both spatial and frequency domains, enabling more comprehensive scene recovery. As a result, our algorithm achieves more balanced improvements in color fidelity, contrast, and structural detail preservation, producing more visually natural results than competing approaches.

In summary, across four challenging restoration tasks, including real-world haze, sandstorm, underwater, and remote sensing scenarios, the proposed SFP consistently produces more visually natural and faithful restoration results than existing methods.

\subsection{Ablation Studies}
We conduct ablation studies to evaluate the effectiveness of each component in the proposed framework.

\begin{figure}[!t]
 \setlength{\abovecaptionskip}{2pt}   
    \setlength{\belowcaptionskip}{-13pt}  
    \centering    
    \includegraphics[width=3.5in]{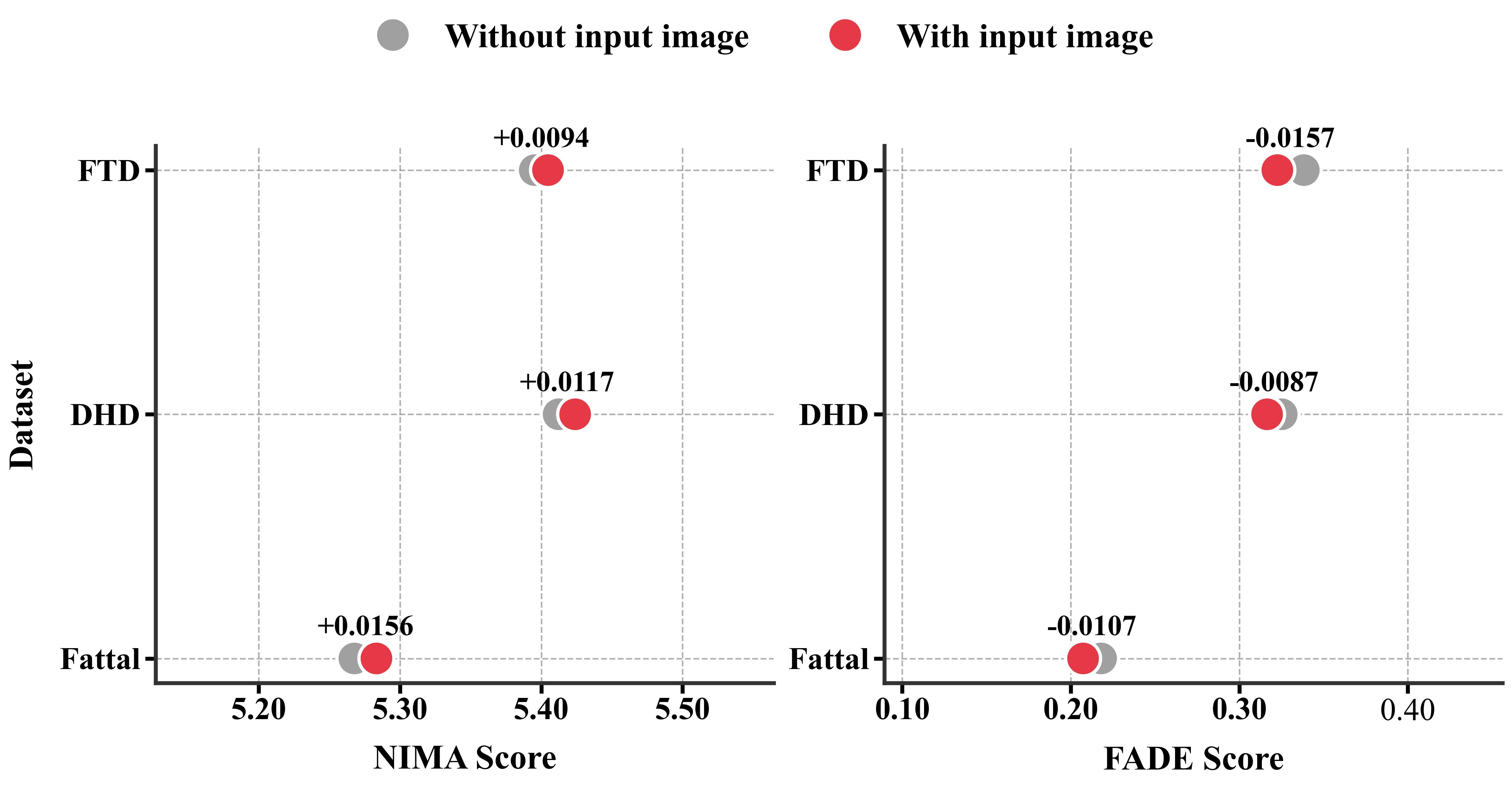}
    \caption{Objective comparison for the ablation of input image fusion. }
    \label{fig:ablation_fusion}
\end{figure}

\textbf{Effectiveness of SDP.} The SDP is designed to estimate an accurate transmission map for scene recovery by inverting the ASM, thereby effectively mitigating scattering-induced degradations. As illustrated in Fig.~\ref{fig:ablation}, removing the SDP component leads to noticeable residual haze, with prominent scattering artifacts remaining in the recovered images.
Quantitatively, as reported in Table~\ref{tab:ablation}, removing SDP results in consistent performance degradation across four real-world datasets, as evidenced by lower NIMA and higher FADE scores. Specifically, the average NIMA score decreases from 5.3578 to 5.1945 (a drop of approximately 3.05\%), while the FADE score increases from 0.3100 to 0.7500 (an increase of approximately 141.94\%). Given that higher NIMA and lower FADE indicate better perceptual quality, these changes clearly reflect degraded visual quality, with increased haze density and more pronounced scattering effects.
Both visual and quantitative results consistently demonstrate that removing SDP exacerbates scattering-induced degradations, highlighting its critical role in effective scene recovery.


\textbf{Effectiveness of FDP.} The FDP is proposed to construct an adaptive mask for frequency enhancement, which effectively compensates for frequency information loss caused by degradations. As illustrated in Fig.~\ref{fig:ablation}, removing the FDP leads to noticeable color distortions, with evident color shifts in the zoomed-in regions. This is because, without FDP, the model fails to adequately recover high-frequency components related to chromatic details, resulting in inaccurate color reconstruction and reduced color fidelity.
Moreover, as reported in Table~\ref{tab:ablation}, removing the FDP leads to a consistent decline in performance across four real-world datasets (Fattal, FTD, HSTS, and DHQ), as indicated by lower NIMA and higher FADE scores. Specifically, the average NIMA score decreases from 5.3578 to 5.1311 (a drop of approximately 4.23\%), and the FADE score increases from 0.3100 to 0.6122 (an increase of approximately 97.48\%). Given that higher NIMA and lower FADE correspond to better perceptual quality, these changes indicate degraded visual quality, particularly in terms of color consistency and naturalness.
Both visual and quantitative results demonstrate that the FDP plays a crucial role in preserving high-frequency details and maintaining faithful color reconstruction.


\begin{figure*}[!t]
    \centering
    \setlength{\abovecaptionskip}{4pt}  
    \setlength{\belowcaptionskip}{-10pt}  
    \includegraphics[width=7.15in]{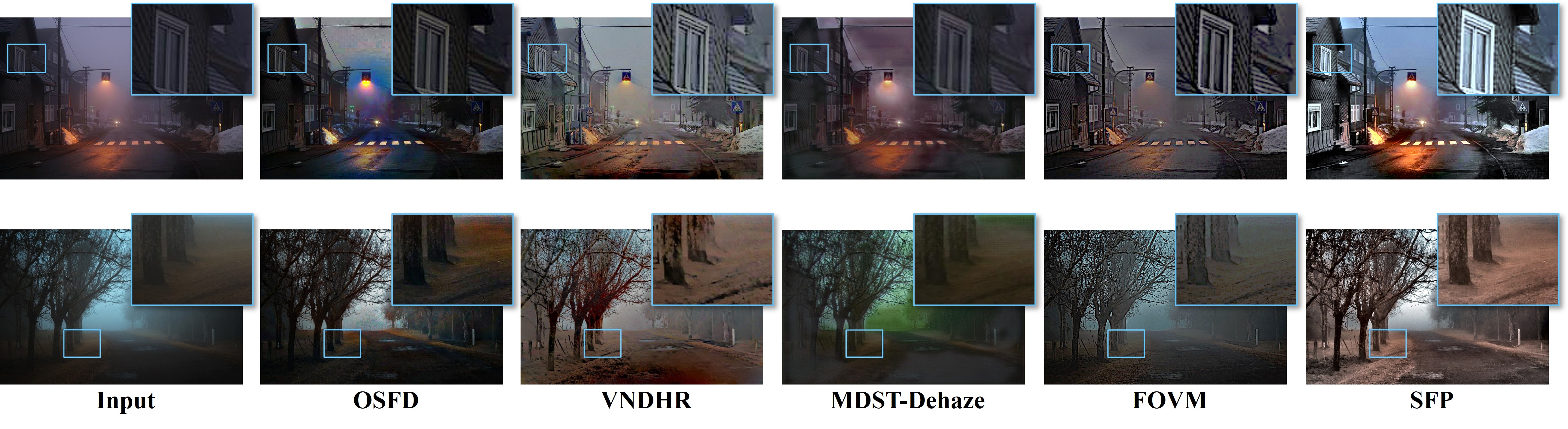}
    \caption{Comparison of state-of-the-art methods on real-world nighttime hazy images.} 
    \label{fig:nighttime}
\end{figure*}
\begin{figure*}[!t]
    \centering
    \setlength{\abovecaptionskip}{4pt}  
    \setlength{\belowcaptionskip}{-10pt}  
    \includegraphics[width=7.15in]{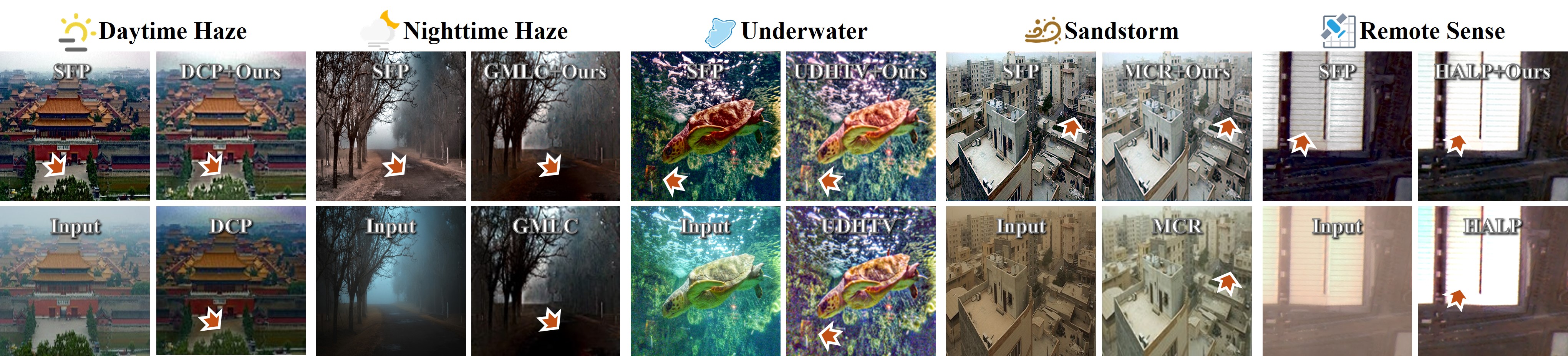}
    \caption{Plug-and-play integration of our algorithm with existing spatial-domain techniques for performance improvement.} 
    \label{fig:ablation_our_other}
\end{figure*}

\textbf{Effectiveness of Fusion.} To assess the effectiveness of the proposed fusion strategy, we compare it with a naive alternative that directly averages the pixel values of the intermediate results. As shown in Fig.~\ref{fig:ablation}, this naive scheme (denoted as “w/o fusion”) fails to properly exploit the complementary characteristics of SDP and FDP, leading to over-smoothed details, reduced contrast, and less faithful color representation. Quantitative results in Table~\ref{tab:ablation} further support this observation. Compared with the proposed fusion strategy, the naive fusion results in a decrease in the average NIMA score from 5.3578 to 5.2896 (a drop of approximately 1.27\%) and an increase in the FADE score from 0.3100 to 0.6698 (an increase of approximately 116.06\%). These degradations confirm that simple averaging is insufficient for effectively combining the two priors.
In contrast, the proposed weighted fusion strategy adaptively balances spatial and frequency information, enabling more effective integration of their respective strengths and producing results with sharper details, improved contrast, and more natural color appearance.

Furthermore, to quantitatively demonstrate the effectiveness of fusing the input image into the proposed fusion framework, we conduct objective comparisons on four datasets, namely Fattal, HSTS, FTD, and DHQ. Fig.~\ref{fig:ablation_fusion} presents the average scores obtained with and without fusing the input image across the four datasets. It can be observed that fusing the input image effectively improves the NIMA score while reducing the FADE value, leading to more natural and visually pleasing restoration results.


\textbf{Effectiveness of Post-processing.} The post-processing (PP) component consists of gamma correction and high dynamic range (HDR) compression, in which the gamma correction adjusts the overall brightness to achieve more perceptually balanced illumination, and HDR compression suppresses over-exposed regions and preserves details in high-intensity areas, leading to improved visual consistency.
As illustrated in Fig.~\ref{fig:ablation}, incorporating the PP module produces visually more pleasing results, with better contrast and more natural brightness compared to those without post-processing.
Quantitative results in Table~\ref{tab:ablation} further validate its effectiveness. With the inclusion of PP, the average NIMA score on four real-world datasets increases from 5.2930 to 5.3578 (an improvement of approximately 1.22\%), while the FADE score decreases from 0.3102 to 0.3100 (a reduction of approximately 0.06\%). These improvements demonstrate that the PP component effectively enhances visual quality, particularly in terms of luminance balance and contrast refinement.
Both visual and quantitative comparisons confirm that the proposed post-processing step consistently improves the final recovered results.


\begin{figure*}[!t]
    \centering
    \setlength{\abovecaptionskip}{4pt}  
    \setlength{\belowcaptionskip}{-10pt}  
    \includegraphics[width=7.15in]{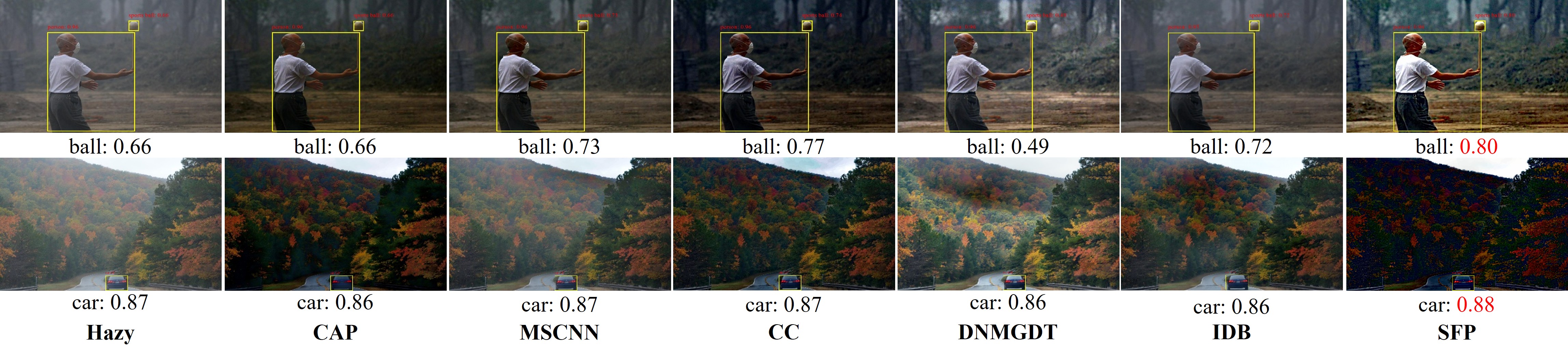}
    \caption{Object recognition results on real-world hazy and dehazed images. } 
    \label{fig:objection}
\end{figure*}

\begin{figure}[!t]
    \centering
    \setlength{\abovecaptionskip}{4pt}  
    \setlength{\belowcaptionskip}{-10pt}  
    \includegraphics[width=3.5in]{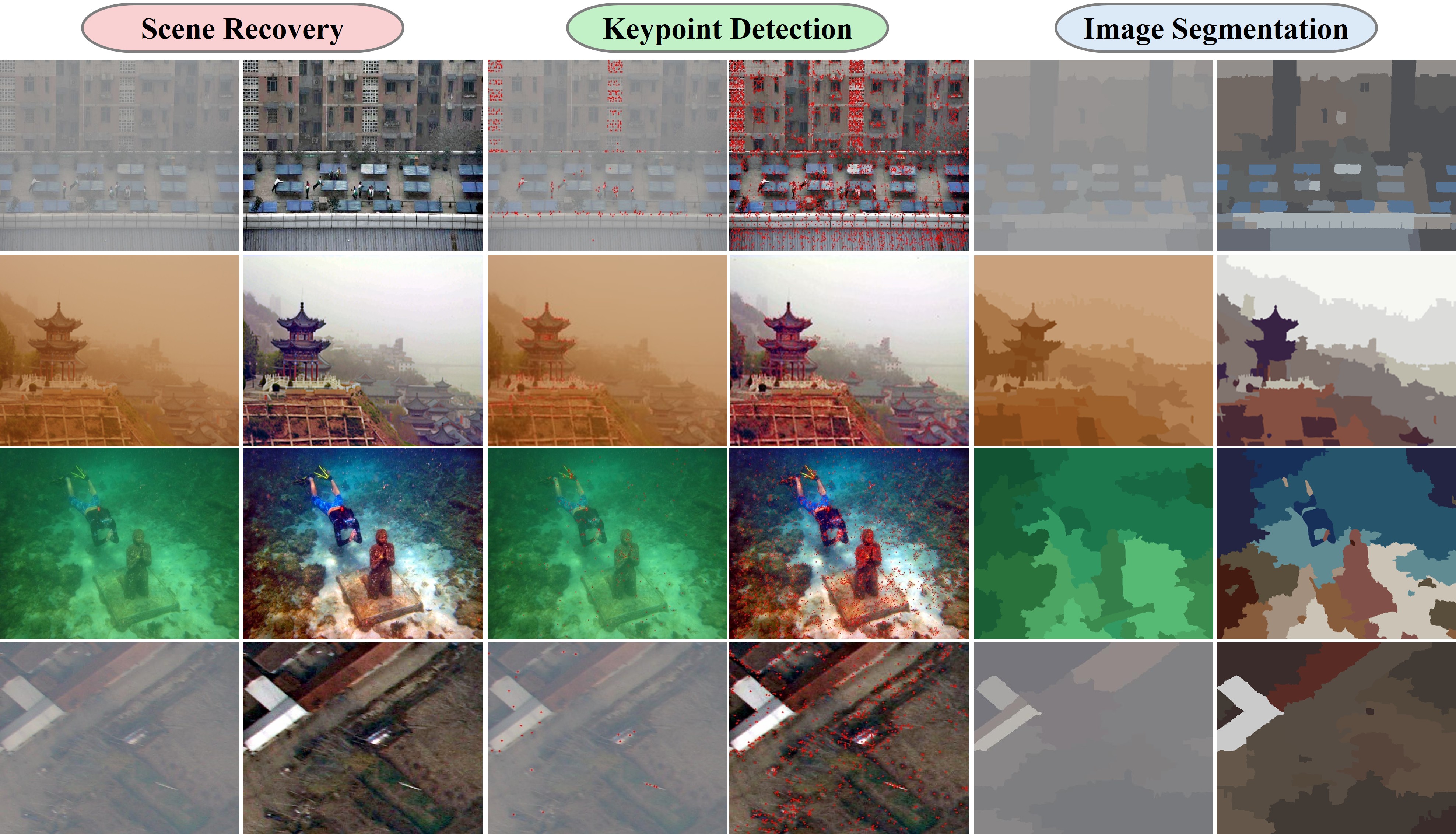}
    \caption{Visual comparison results on high-level vision tasks, including keypoint detection and image segmentation, for degraded images and their corresponding recovered results. } 
    \label{fig:high_level}
\end{figure}

\subsection{Generalization to Nighttime Hazy Scenes}
To further evaluate the generalization capability of the proposed framework, we conduct experiments on real-world nighttime hazy images with multiple coupled degradations. As is well known, nighttime hazy scenes usually suffer from complex degradation factors, including haze scattering, low illumination, color distortion, glow effects, and other coupled degradations, making scene recovery significantly more challenging than other scattering-degraded scenarios. Specifically, we incorporate the scene transmission and atmospheric light estimated by the proposed SDP into an enhanced atmospheric scattering model~\cite{ju2021ide} to address non-uniform illumination. Quantitative results in Table~\ref{table:nighthaze} report comparisons on two real-world nighttime haze datasets, namely NHRW and RealNightHaze, against several state-of-the-art methods. Specifically, the proposed method improves the average NIMA score from 4.9215 to 5.0912 (an improvement of approximately 3.45\%), while reducing the FADE score from 0.4003 to 0.3141 (a relative reduction of approximately 21.53\%). These results demonstrate that the proposed method maintains strong scene recovery capability even under complex coupled degradation conditions. The visual comparisons in Fig.~\ref{fig:nighttime} further validate the superiority of the proposed method on real-world nighttime hazy images. Compared with existing specialized approaches, our SFP not only effectively removes nighttime haze to recover clearer scene structures, but also improves illumination and restores more faithful color representation. Overall, the proposed algorithm yields higher-quality and more visually coherent recovery results under complex nighttime coupled degradation conditions, highlighting its strong generalization ability and robustness in real-world scene recovery tasks.


\begin{table}[!t]
	\renewcommand\arraystretch{1.2}
	\setlength{\tabcolsep}{3pt}
	\centering
	\caption{Quantitative comparison of the proposed SFP and state-of-the-art methods on two real-world nighttime hazy image datasets.}
	\vspace{-0.2cm}
	\begin{tabular}{c|cc|cc|cc}
		\toprule
		\multirow{2}{*}{Methods}
		& \multicolumn{2}{c|}{NHRW}
		& \multicolumn{2}{c|}{RealNightHaze}
		& \multicolumn{2}{c}{Average} \\
		\cmidrule{2-7}
		~
		& NIMA\textcolor{cyan!60!blue}{$\uparrow$} 
		& FADE\textcolor{cyan!60!blue}{$\downarrow$}
		& NIMA\textcolor{cyan!60!blue}{$\uparrow$} 
		& FADE\textcolor{cyan!60!blue}{$\downarrow$}
		& NIMA\textcolor{cyan!60!blue}{$\uparrow$} 
		& FADE\textcolor{cyan!60!blue}{$\downarrow$} \\
		\midrule

		MRP~\cite{zhang2017fast}
		& 4.7456 & 0.4414 & 4.9348 & 0.4695 
		& 4.8402 & 0.4555 \\

		FastMRP~\cite{zhang2017fast}
		& 4.5458 & 0.3886 & 4.6863 & \cellcolor{second}\underline{0.4119} 
		& 4.6161 & \cellcolor{second}\underline{0.4003} \\

		OSFD~\cite{zhang2020nighttime}
		& \cellcolor{second}\underline{4.8570} & \cellcolor{second}\underline{0.3804} & 4.9860 & 0.4193 
		& \cellcolor{second}\underline{4.9215} & 0.3999 \\

		UVRM~\cite{liu2023multi}
		& 4.7484 & 0.4544 & 4.9577 & 0.4870 
		& 4.8531 & 0.4707 \\

		VNDHR~\cite{liu2025vndhr}
		& 4.7982 & 0.4910 & 4.9299 & 0.5502 
		& 4.8641 & 0.5206 \\

		CoA~\cite{ma2025coa}
		& 4.7494 & 0.6826 & \cellcolor{second}\underline{5.0804} & 0.7551 
		& 4.9149 & 0.7189 \\

		MDST-Dehaze~\cite{huang2026haze}
		& 4.3842 & 0.6466 & 4.6505 & 0.7670 
		& 4.5174 & 0.7068 \\
        
        FOVM~\cite{liu2026fovm} 
        & 4.8482 & 0.3951 & 4.9838 & 0.4518 
        & 4.9160 & 0.4235 \\

		SFP
		& \cellcolor{best}\textbf{5.0551} & \cellcolor{best}\textbf{0.3066} & \cellcolor{best}\textbf{5.1273} & \cellcolor{best}\textbf{0.3216} 
		& \cellcolor{best}\textbf{5.0912} & \cellcolor{best}\textbf{0.3141} \\

		\bottomrule
	\end{tabular}
	\vspace{-0.2cm}
	\label{table:nighthaze}
\end{table}
\subsection{Plug-and-Play Property} 
As illustrated in Fig.~\ref{fig:ablation_our_other}, the proposed framework exhibits a flexible plug-and-play capability, where the spatial-domain prior (SDP) can be readily replaced by a variety of existing spatial priors, including DCP~\cite{he2011single}, GMLC~\cite{li2015nighttime}, UDHTV~\cite{li2025dual}, MCR~\cite{al2024increasing}, and HALP~\cite{he2023remote}. By incorporating our frequency-domain prior (FDP), these spatial-domain methods consistently achieve improved scene recovery performance, yielding enhanced visual quality in terms of contrast, detail preservation, and color fidelity. 
This observation indicates that the proposed FDP is not constrained to any specific spatial prior and can be seamlessly embedded into a wide range of spatial-domain restoration pipelines. The consistent performance gains further demonstrate that FDP effectively complements spatial priors by compensating for their limitations in frequency modeling, particularly in recovering high-frequency components that are often degraded or suppressed in scattering scenarios.
Moreover, the proposed SFP consistently outperforms all alternative combinations, highlighting not only the effectiveness of the designed spatial prior but also the benefit of jointly exploiting spatial and frequency information. Overall, these results validate the flexibility of the proposed plug-and-play framework and support the effectiveness of and transferability of the frequency-domain prior.


\begin{table}[!t]
    \renewcommand\arraystretch{1.2}
    \setlength{\tabcolsep}{3pt}
    \centering
    \caption{Performance Comparison with Average Scores}
    \begin{tabular}{c|ccccc|c}
    \toprule
        Method & Person & Bicycle & Car & Motorbike & Bus & Average \\ \midrule
        Hazy   & 0.9264 & 0.8942 & 0.8858 & 0.9116 & 0.7569 & 0.8750 \\ 
        CAP    & 0.9287 & 0.8009 & 0.8726 & 0.8625 & 0.8663 & 0.8662 \\
        MSCNN  & \cellcolor{second}\underline{0.9346} & 0.8544 & 0.8955 & 0.9119 & 0.9084 & 0.9010 \\ 
        MR     & 0.9254 & 0.9027 & 0.8685 & 0.8721 & 0.7788 & 0.8695 \\ 
        GCP    & 0.8581 & 0.8320 & 0.6996 & 0.5325 & 0.6139 & 0.7072 \\ 
        SBTE   & 0.9071 & 0.8295 & 0.8598 & 0.7544 & \cellcolor{second}\underline{0.9362} & 0.8574 \\
        CC     & 0.9334 & 0.8978 & \cellcolor{second}\underline{0.8967} & 0.9067 & 0.9140 & 0.9097 \\ 
        ROP    & 0.9250 & 0.8823 & 0.8925 & 0.8845 & 0.9168 & 0.9002 \\ 
        ROP+   & 0.9065 & 0.8040 & 0.8565 & 0.7018 & 0.7620 & 0.8062 \\ 
        COA    & 0.9308 & 0.8810 & 0.8920 & 0.8746 & 0.9168 & 0.8990 \\ 
        DNMGDT & \cellcolor{second}\underline{0.9346} & 0.8722 & 0.8909 & 0.8752 & 0.9084 & 0.8963 \\
        IDB    & 0.8810 & 0.8618 & 0.7830 & 0.7449 & 0.6022 & 0.7746 \\ 
        SDO    & 0.9314 & \cellcolor{second}\underline{0.9048} & 0.8954 & \cellcolor{second}\underline{0.9390} & 0.9309 & \cellcolor{second}\underline{0.9203} \\ 
        PDDA   & 0.9318 & 0.8726 & 0.8747 & 0.9152 & 0.9168 & 0.9022 \\ 
        SFP    & \cellcolor{best}\textbf{0.9408} & \cellcolor{best}\textbf{0.9215} & \cellcolor{best}\textbf{0.9074} & \cellcolor{best}\textbf{0.9416} & \cellcolor{best}\textbf{0.9693} & \cellcolor{best}\textbf{0.9361} \\ \bottomrule
    \end{tabular}
    \vspace{-0.2cm}
    \label{tab:high_level}
\end{table}
\subsection{Performance Improvements on High-Level Vision Tasks}
The proposed SFP not only effectively recovers scattering-degraded images but also significantly improves the performance of downstream high-level computer vision tasks. To demonstrate the superiority of the proposed method in boosting task performance, we employ YOLOv8~\cite{varghese2024yolov8} as the benchmark model to evaluate the effectiveness of SFP against several competitive dehazing methods for object detection, as illustrated in Fig.~\ref{fig:objection}. It can be observed that our algorithm achieves the highest detection accuracy among all compared approaches. Furthermore, we randomly select 500 hazy images from the RTTS dataset~\cite{li2019benchmarking} and quantitatively evaluate the proposed method using average precision as the evaluation metric, with the results summarized in Table~\ref{tab:high_level}. As reported in Table~\ref{tab:high_level}, the proposed approach achieves the best performance across all five categories. 
From visual comparisons, Fig.~\ref{fig:high_level} presents qualitative results on additional high-level vision tasks, including keypoint detection~\cite{lowe2004distinctive} and image segmentation~\cite{lei2018superpixel}. Overall, these results demonstrate that the proposed algorithm effectively enhances downstream high-level vision tasks by providing higher-quality images.

\section{Conclusion}
In this paper, we have presented a unified scene recovery framework based on spatial and frequency priors for addressing diverse real-world scattering degradations. The proposed spatial-domain prior (SDP) estimates scene transmission by modeling the projection relationship between the inverted degraded image and its spectral direction, thereby alleviating scattering-induced contrast degradation in the spatial domain. Meanwhile, the proposed frequency-domain priors (FDP) enhance image brightness, color fidelity, and high-frequency details through adaptive frequency modulation in the Fourier domain. Finally, a weighted fusion strategy is introduced to effectively combine the complementary advantages of spatial- and frequency-domain recovered results for high-quality scene recovery.
Unlike existing prior- and learning-based methods, the proposed SFP jointly exploits complementary degradation characteristics in both spatial and frequency domains without relying on large-scale synthetic training data. Experiments demonstrate that the proposed SFP consistently outperforms state-of-the-art approaches across various real-world scattering-degraded scenarios, including haze, sandstorm, underwater, and remote sensing scenes. Furthermore, our SFP also remains effective in complex real-world nighttime hazy scenes involving multiple coupled degradations and benefits downstream high-level vision tasks.

\bibliographystyle{ieeetr}

\begin{thebibliography}{10}

\bibitem{he2011single}
K.~He, J.~Sun, and X.~Tang, ``Single image haze removal using dark channel prior,'' {\em IEEE Transactions on Pattern Analysis and Machine Intelligence}, vol.~33, no.~12, pp.~2341--2353, 2011.

\bibitem{zhu2015a}
Q.~Zhu, J.~Mai, and L.~Shao, ``A fast single image haze removal algorithm using color attenuation prior,'' {\em IEEE Transactions on Image Processing}, vol.~24, no.~11, pp.~3522--3533, 2015.

\bibitem{li2016underwater}
C.-Y. Li, J.-C. Guo, R.-M. Cong, Y.-W. Pang, and B.~Wang, ``Underwater image enhancement by dehazing with minimum information loss and histogram distribution prior,'' {\em IEEE Trans. Image Process.}, vol.~25, no.~12, pp.~5664--5677, 2016.

\bibitem{zhang2017fast}
J.~Zhang, Y.~Cao, S.~Fang, Y.~Kang, and C.~Wen~Chen, ``Fast haze removal for nighttime image using maximum reflectance prior,'' in {\em Proceedings of the IEEE conference on computer vision and pattern recognition}, pp.~7418--7426, 2017.

\bibitem{ju2019idgcp}
M.~Ju, C.~Ding, Y.~J. Guo, and D.~Zhang, ``{IDGCP}: Image dehazing based on gamma correction prior,'' {\em IEEE Trans. Image Process.}, vol.~29, pp.~3104--3118, 2019.

\bibitem{berman2020single}
D.~Berman, T.~Treibitz, and S.~Avidan, ``Single image dehazing using haze-lines,'' {\em IEEE Transactions on Pattern Analysis and Machine Intelligence}, vol.~42, no.~3, pp.~720--734, 2020.

\bibitem{ju2021idrlp}
M.~Ju, C.~Ding, C.~A. Guo, W.~Ren, and D.~Tao, ``{IDRLP}: Image dehazing using region line prior,'' {\em IEEE Transactions on Image Processing}, vol.~30, pp.~9043--9057, 2021.

\bibitem{zhuang2022underwater}
P.~Zhuang, J.~Wu, F.~Porikli, and C.~Li, ``Underwater image enhancement with hyper-laplacian reflectance priors,'' {\em IEEE Trans. Image Process.}, vol.~31, pp.~5442--5455, 2022.

\bibitem{ling2023single}
P.~Ling, H.~Chen, X.~Tan, Y.~Jin, and E.~Chen, ``Single image dehazing using saturation line prior,'' {\em IEEE Transactions on Image Processing}, vol.~32, pp.~3238--3253, 2023.

\bibitem{wu2026image}
X.~Wu, T.~Lyu, and M.~Ju, ``Image dehazing using patch-wise nonlinear brightness prior,'' {\em IEEE Signal Processing Letters}, vol.~33, pp.~1140--1144, 2026.

\bibitem{liu2026ihdcp}
Y.~Liu, T.~Li, C.~Tan, W.~Ren, C.~Ancuti, and W.~Lin, ``Ihdcp: Single image dehazing using inverted haze density correction prior,'' {\em IEEE Transactions on Image Processing}, 2026.

\bibitem{peng2018generalization}
Y.-T. Peng, K.~Cao, and P.~C. Cosman, ``Generalization of the dark channel prior for single image restoration,'' {\em IEEE Transactions on Image Processing}, vol.~27, no.~6, pp.~2856--2868, 2018.

\bibitem{liu2021rank}
J.~Liu, W.~Liu, J.~Sun, and T.~Zeng, ``Rank-one prior: Toward real-time scene recovery,'' in {\em Proceedings of the IEEE/CVF conference on computer vision and pattern recognition}, pp.~14802--14810, 2021.

\bibitem{liu2023rank}
J.~Liu, R.~W. Liu, J.~Sun, and T.~Zeng, ``{Rank-One Prior:} real-time scene recovery,'' {\em IEEE Transactions on Pattern Analysis and Machine Intelligence}, vol.~45, no.~7, pp.~8845--8860, 2023.

\bibitem{he2025alsp}
L.~He, Z.~Yi, J.~Liu, C.~Chen, M.~Lu, and Z.~Chen, ``Alsp+: fast scene recovery via ambient light similarity prior,'' {\em IEEE Trans. Image Process.}, vol.~34, pp.~4470--4484, 2025.

\bibitem{cai}
B.~Cai, X.~Xu, K.~Jia, C.~Qing, and D.~Tao, ``{DehazeNet}: An end-to-end system for single image haze removal,'' {\em IEEE Trans. Image Process.}, vol.~25, no.~11, pp.~5187--5198, 2016.

\bibitem{valanarasu2022transweather}
J.~M.~J. Valanarasu, R.~Yasarla, and V.~M. Patel, ``Transweather: Transformer-based restoration of images degraded by adverse weather conditions,'' in {\em CVPR}, pp.~2353--2363, 2022.

\bibitem{gao2023let}
Y.~Gao, W.~Xu, and Y.~Lu, ``Let you see in haze and sandstorm: Two-in-one low-visibility enhancement network,'' {\em IEEE Transactions on Instrumentation and Measurement}, vol.~72, pp.~1--12, 2023.

\bibitem{wen2023encoder}
Y.~Wen, T.~Gao, J.~Zhang, Z.~Li, and T.~Chen, ``Encoder-free multiaxis physics-aware fusion network for remote sensing image dehazing,'' {\em IEEE Transactions on Geoscience and Remote Sensing}, vol.~61, pp.~1--15, 2023.

\bibitem{feng2024advancing}
Y.~Feng, L.~Ma, X.~Meng, F.~Zhou, R.~Liu, and Z.~Su, ``Advancing real-world image dehazing: Perspective, modules, and training,'' {\em IEEE Transactions on Pattern Analysis and Machine Intelligence}, vol.~46, no.~12, pp.~9303--9320, 2024.

\bibitem{cong2024semi}
X.~Cong, J.~Gui, J.~Zhang, J.~Hou, and H.~Shen, ``A semi-supervised nighttime dehazing baseline with spatial-frequency aware and realistic brightness constraint,'' in {\em Proceedings of the IEEE/CVF Conference on Computer Vision and Pattern Recognition}, pp.~2631--2640, 2024.

\bibitem{lu2024aosrnet}
Y.~Lu, D.~Yang, Y.~Gao, R.~W. Liu, J.~Liu, and Y.~Guo, ``Aosrnet: All-in-one scene recovery networks via multi-knowledge integration,'' {\em Knowledge-Based Systems}, vol.~294, p.~111786, 2024.

\bibitem{su2025real}
Y.~Su, N.~Wang, Z.~Cui, Y.~Cai, C.~He, and A.~Li, ``Real scene single image dehazing network with multi-prior guidance and domain transfer,'' {\em IEEE Transactions on Multimedia}, 2025.

\bibitem{fu2025iterative}
J.~Fu, S.~Liu, Z.~Liu, C.-L. Guo, H.~Park, R.~Wu, G.~Wang, and C.~Li, ``Iterative predictor-critic code decoding for real-world image dehazing,'' in {\em CVPR}, pp.~12700--12709, 2025.

\bibitem{ning2025mabdt}
J.~Ning, J.~Yin, F.~Deng, and L.~Xie, ``Mabdt: Multi-scale attention boosted deformable transformer for remote sensing image dehazing,'' {\em Signal Processing}, vol.~229, p.~109768, 2025.

\bibitem{liu2024real}
R.~W. Liu, Y.~Lu, Y.~Gao, Y.~Guo, W.~Ren, F.~Zhu, and F.-Y. Wang, ``Real-time multi-scene visibility enhancement for promoting navigational safety of vessels under complex weather conditions,'' {\em IEEE Transactions on Intelligent Transportation Systems}, 2024.

\bibitem{li2025low}
Z.~Li, W.~Kuang, B.~Bhanu, Y.~Deng, Y.~Chen, and K.~Xu, ``Low-visibility scene enhancement by isomorphic dual-branch framework with attention learning,'' {\em IEEE Transactions on Intelligent Transportation Systems}, vol.~26, no.~5, pp.~7127--7141, 2025.

\bibitem{wen2025multi}
Y.~Wen, T.~Gao, J.~Zhang, Z.~Li, and T.~Chen, ``Multi-axis prompt and multi-dimension fusion network for all-in-one weather-degraded image restoration,'' in {\em Proceedings of the AAAI Conference on Artificial Intelligence}, vol.~39, pp.~8323--8331, 2025.

\bibitem{cui2025adair}
Y.~Cui, S.~W. Zamir, S.~Khan, A.~Knoll, M.~Shah, and F.~S. Khan, ``Adair: Adaptive all-in-one image restoration via frequency mining and modulation,'' in {\em ICLR}, pp.~57335--57356, 2025.

\bibitem{Chen_2025_CVPR}
I.-H. Chen, W.-T. Chen, Y.-W. Liu, Y.-C. Chiang, S.-Y. Kuo, and M.-H. Yang, ``Unirestore: Unified perceptual and task-oriented image restoration model using diffusion prior,'' in {\em CVPR}, pp.~17969--17979, June 2025.

\bibitem{ma2025coa}
L.~Ma, Y.~Feng, Y.~Zhang, J.~Liu, W.~Wang, G.-Y. Chen, C.~Xu, and Z.~Su, ``Coa: Towards real image dehazing via compression-and-adaptation,'' in {\em Proceedings of the Computer Vision and Pattern Recognition Conference}, pp.~11197--11206, 2025.

\bibitem{zhang2026pdda}
Y.~Zhang, Y.~Feng, X.~Li, F.~Zhou, and Z.~Su, ``Pdda: Prompt-driven domain adaptation for real-world image dehazing,'' {\em IEEE Transactions on Image Processing}, 2026.

\bibitem{wang2026sea}
J.~Wang, H.~Bi, J.~Cao, F.~Gao, and J.~Dong, ``Sea-pace: Semi-supervised underwater image enhancement via gaussian process--assisted self-paced learning,'' in {\em Proceedings of the AAAI Conference on Artificial Intelligence}, vol.~40, pp.~9930--9938, 2026.

\bibitem{lin2026downstream}
B.~Lin, F.~Gao, Y.~Yu, J.~Dong, and Q.~Du, ``Downstream task inspired underwater image enhancement: A perception-aware study from dataset construction to network design,'' {\em IEEE Transactions on Image Processing}, 2026.

\bibitem{gui2026brightness}
J.~Gui, X.~Cong, Y.-X. Zhang, J.~Hou, and D.~Tao, ``Brightness-aware synthetic-to-real learning for nighttime hazy image enhancement,'' {\em IEEE Transactions on Pattern Analysis and Machine Intelligence}, 2026.

\bibitem{cui2026starir}
Y.~Cui, S.~W. Zamir, M.-H. Yang, A.~Knoll, F.~S. Khan, and S.~Khan, ``Starir: Convolutional image restoration with spatial-frequency fusion,'' {\em IEEE Transactions on Pattern Analysis and Machine Intelligence}, pp.~1--18, 2026.

\bibitem{liu2023multi}
Y.~Liu, Z.~Yan, J.~Tan, and Y.~Li, ``Multi-purpose oriented single nighttime image haze removal based on unified variational retinex model,'' {\em IEEE Transactions on Circuits and Systems for Video Technology}, vol.~33, no.~4, pp.~1643--1657, 2023.

\bibitem{liu2025vndhr}
Y.~Liu, X.~Wang, E.~Hu, A.~Wang, B.~Shiri, and W.~Lin, ``{VNDHR}: Variational single nighttime image dehazing for enhancing visibility in intelligent transportation systems via hybrid regularization,'' {\em IEEE Transactions on Intelligent Transportation Systems}, pp.~1--15, 2025.

\bibitem{liu2026fovm}
Y.~Liu, T.~Li, Z.~Zhou, W.~Ren, and W.~Lin, ``Real-world nighttime image dehazing via bayesian-based fractional-order variational model,'' {\em IEEE Transactions on Image Processing}, pp.~1--13, 2026.

\bibitem{ren2016single}
W.~Ren, S.~Liu, H.~Zhang, J.~Pan, X.~Cao, and M.-H. Yang, ``Single image dehazing via multi-scale convolutional neural networks,'' in {\em Computer Vision -- ECCV 2016} (B.~Leibe, J.~Matas, N.~Sebe, and M.~Welling, eds.), (Cham), pp.~154--169, Springer International Publishing, 2016.

\bibitem{narasimhan2002vision}
S.~G. Narasimhan and S.~K. Nayar, ``Vision and the atmosphere,'' {\em International journal of computer vision}, vol.~48, no.~3, pp.~233--254, 2002.

\bibitem{liu2021synthetic}
Y.~Liu, L.~Zhu, S.~Pei, H.~Fu, J.~Qin, Q.~Zhang, L.~Wan, and W.~Feng, ``From synthetic to real: Image dehazing collaborating with unlabeled real data,'' in {\em Proceedings of the 29th ACM international conference on multimedia}, pp.~50--58, 2021.

\bibitem{li2019underwater}
C.~Li, C.~Guo, W.~Ren, R.~Cong, J.~Hou, S.~Kwong, and D.~Tao, ``An underwater image enhancement benchmark dataset and beyond,'' {\em IEEE Trans. Image Process.}, vol.~29, pp.~4376--4389, 2019.

\bibitem{zhu2025real}
Z.-H. Zhu, W.~Lu, S.-B. Chen, C.~H. Ding, J.~Tang, and B.~Luo, ``Real-world remote sensing image dehazing: Benchmark and baseline,'' {\em IEEE Transactions on Geoscience and Remote Sensing}, 2025.

\bibitem{he2013}
K.~He, J.~Sun, and X.~Tang, ``Guided image filtering,'' {\em IEEE Trans. Pattern Anal. Mach. Intell.}, vol.~35, no.~6, pp.~1397--1409, 2013.

\bibitem{meng2013efficient}
G.~Meng, Y.~Wang, J.~Duan, S.~Xiang, and C.~Pan, ``Efficient image dehazing with boundary constraint and contextual regularization,'' in {\em 2013 IEEE International Conference on Computer Vision}, pp.~617--624, 2013.

\bibitem{salazar2018fast}
S.~Salazar-Colores, E.~Cabal-Yepez, J.~M. Ramos-Arreguin, G.~Botella, L.~M. Ledesma-Carrillo, and S.~Ledesma, ``A fast image dehazing algorithm using morphological reconstruction,'' {\em IEEE transactions on image processing}, vol.~28, no.~5, pp.~2357--2366, 2018.

\bibitem{ju2020idgcp}
M.~Ju, C.~Ding, Y.~J. Guo, and D.~Zhang, ``{IDGCP}: Image dehazing based on gamma correction prior,'' {\em IEEE Transactions on Image Processing}, vol.~29, pp.~3104--3118, 2020.

\bibitem{kim2020fast}
S.~E. Kim, T.~H. Park, and I.~K. Eom, ``Fast single image dehazing using saturation based transmission map estimation,'' {\em IEEE Transactions on Image Processing}, vol.~29, pp.~1985--1998, 2020.

\bibitem{dhara2020color}
S.~K. Dhara, M.~Roy, D.~Sen, and P.~K. Biswas, ``Color cast dependent image dehazing via adaptive airlight refinement and non-linear color balancing,'' {\em IEEE Transactions on Circuits and Systems for Video Technology}, vol.~31, no.~5, pp.~2076--2081, 2020.

\bibitem{ling2025efficient}
P.~Ling, H.~Chen, H.~Wang, Y.~Gu, Y.~Jin, J.~Zheng, and E.~Chen, ``Efficient haze removal via scene depth ordering for robust traffic monitoring,'' {\em IEEE Transactions on Intelligent Transportation Systems}, pp.~1--14, 2025.

\bibitem{fu2014fusion}
X.~Fu, Y.~Huang, D.~Zeng, X.-P. Zhang, and X.~Ding, ``A fusion-based enhancing approach for single sandstorm image,'' in {\em 2014 IEEE 16th international workshop on multimedia signal processing (MMSP)}, pp.~1--5, IEEE, 2014.

\bibitem{al2016visibility}
Z.~Al-Ameen, ``Visibility enhancement for images captured in dusty weather via tuned tri-threshold fuzzy intensification operators,'' {\em International Journal of Intelligent Systems and Applications}, vol.~8, no.~8, p.~10, 2016.

\bibitem{peng2019image}
Y.-T. Peng, Z.~Lu, F.-C. Cheng, Y.~Zheng, and S.-C. Huang, ``Image haze removal using airlight white correction, local light filter, and aerial perspective prior,'' {\em IEEE Transactions on Circuits and Systems for Video Technology}, vol.~30, no.~5, pp.~1385--1395, 2019.

\bibitem{shi2019let}
Z.~Shi, Y.~Feng, M.~Zhao, E.~Zhang, and L.~He, ``Let you see in sand dust weather: A method based on halo-reduced dark channel prior dehazing for sand-dust image enhancement,'' {\em Ieee Access}, vol.~7, pp.~116722--116733, 2019.

\bibitem{jeon2022sand}
J.-J. Jeon, T.-H. Park, and I.-K. Eom, ``Sand-dust image enhancement using chromatic variance consistency and gamma correction-based dehazing,'' {\em Sensors}, vol.~22, no.~23, p.~9048, 2022.

\bibitem{al2024increasing}
Z.~Al-Ameen, ``Increasing the lucidity of sandstorm images using a multistep color reparation algorithm,'' {\em Signal, Image and Video Processing}, vol.~18, no.~11, pp.~8005--8017, 2024.

\bibitem{peng2017underwater}
Y.-T. Peng and P.~C. Cosman, ``Underwater image restoration based on image blurriness and light absorption,'' {\em IEEE transactions on image processing}, vol.~26, no.~4, pp.~1579--1594, 2017.

\bibitem{ancuti2018color}
C.~O. Ancuti, C.~Ancuti, C.~De~Vleeschouwer, and P.~Bekaert, ``Color balance and fusion for underwater image enhancement,'' {\em IEEE Trans. Image Process.}, vol.~27, no.~1, pp.~379--393, 2018.

\bibitem{yuan2021tebcf}
J.~Yuan, Z.~Cai, and W.~Cao, ``Tebcf: Real-world underwater image texture enhancement model based on blurriness and color fusion,'' {\em IEEE Transactions on Geoscience and Remote Sensing}, vol.~60, pp.~1--15, 2021.

\bibitem{zhang2024wwpf}
W.~Zhang, L.~Zhou, P.~Zhuang, G.~Li, X.~Pan, W.~Zhao, and C.~Li, ``Underwater image enhancement via weighted wavelet visual perception fusion,'' {\em IEEE Transactions on Circuits and Systems for Video Technology}, vol.~34, no.~4, pp.~2469--2483, 2024.

\bibitem{zhang2024pcfb}
W.~Zhang, Q.~Liu, Y.~Feng, L.~Cai, and P.~Zhuang, ``Underwater image enhancement via principal component fusion of foreground and background,'' {\em IEEE Transactions on Circuits and Systems for Video Technology}, vol.~34, no.~11, pp.~10930--10943, 2024.

\bibitem{wang2024underwater}
J.~Wang, M.~Wan, Y.~Xu, X.~Kong, G.~Gu, and Q.~Chen, ``Underwater image restoration via constrained color compensation and background light color space-based haze-line model,'' {\em IEEE Transactions on Geoscience and Remote Sensing}, 2024.

\bibitem{zhang2025underwater}
W.~Zhang, M.~Wang, P.~Zhuang, and D.~Liu, ``Underwater image enhancement via advantage feature weighted fusion,'' {\em IEEE Transactions on Circuits and Systems for Video Technology}, 2025.

\bibitem{zhang2025wfac}
W.~Zhang, Q.~Liu, H.~Lu, J.~Wang, and J.~Liang, ``Underwater image enhancement via wavelet decomposition fusion of advantage contrast,'' {\em IEEE Transactions on Circuits and Systems for Video Technology}, 2025.

\bibitem{li2025dual}
Y.~Li, G.~Hou, P.~Zhuang, and Z.~Pan, ``Dual high-order total variation model for underwater image restoration,'' {\em IEEE Transactions on Multimedia}, 2025.

\bibitem{choi2015referenceless}
L.~K. Choi, J.~You, and A.~C. Bovik, ``Referenceless prediction of perceptual fog density and perceptual image defogging,'' {\em IEEE Transactions on Image Processing}, vol.~24, no.~11, pp.~3888--3901, 2015.

\bibitem{min2019objective}
X.~Min, G.~Zhai, K.~Gu, X.~Yang, and X.~Guan, ``Objective quality evaluation of dehazed images,'' {\em IEEE Transactions on Intelligent Transportation Systems}, vol.~20, no.~8, pp.~2879--2892, 2019.

\bibitem{li2019benchmarking}
B.~Li, W.~Ren, D.~Fu, D.~Tao, D.~Feng, W.~Zeng, and Z.~Wang, ``Benchmarking single-image dehazing and beyond,'' {\em IEEE Transactions on Image Processing}, vol.~28, no.~1, pp.~492--505, 2019.

\bibitem{fattal2014dehazing}
R.~Fattal, ``Dehazing using color-lines,'' {\em ACM transactions on graphics (TOG)}, vol.~34, no.~1, pp.~1--14, 2014.

\bibitem{bartani2022adaptive}
A.~Bartani, A.~Abdollahpouri, M.~Ramezani, and F.~A. Tab, ``An adaptive optic-physic based dust removal method using optimized air-light and transfer function,'' {\em Multimedia Tools and Applications}, vol.~81, no.~23, pp.~33823--33849, 2022.

\bibitem{hassaballah2020vehicle}
M.~Hassaballah, M.~A. Kenk, K.~Muhammad, and S.~Minaee, ``Vehicle detection and tracking in adverse weather using a deep learning framework,'' {\em IEEE transactions on intelligent transportation systems}, vol.~22, no.~7, pp.~4230--4242, 2020.

\bibitem{li2019fusion}
H.~Li, J.~Li, and W.~Wang, ``A fusion adversarial underwater image enhancement network with a public test dataset,'' {\em arXiv preprint arXiv:1906.06819}, 2019.

\bibitem{liu2020real}
R.~Liu, X.~Fan, M.~Zhu, M.~Hou, and Z.~Luo, ``Real-world underwater enhancement: Challenges, benchmarks, and solutions under natural light,'' {\em IEEE transactions on circuits and systems for video technology}, vol.~30, no.~12, pp.~4861--4875, 2020.

\bibitem{zheng2023uav}
R.~Zheng and L.~Zhang, ``Uav image haze removal based on saliency-guided parallel learning mechanism,'' {\em IEEE Geoscience and Remote Sensing Letters}, vol.~20, pp.~1--5, 2023.

\bibitem{he2023remote}
Y.~He, C.~Li, and X.~Li, ``Remote sensing image dehazing using heterogeneous atmospheric light prior,'' {\em IEEE Access}, vol.~11, pp.~18805--18820, 2023.

\bibitem{zhang2020nighttime}
J.~Zhang, Y.~Cao, Z.-J. Zha, and D.~Tao, ``Nighttime dehazing with a synthetic benchmark,'' in {\em ACM MM}, pp.~2355--2363, 2020.

\bibitem{jin2023enhancing}
Y.~Jin, B.~Lin, W.~Yan, Y.~Yuan, W.~Ye, and R.~T. Tan, ``Enhancing visibility in nighttime haze images using guided apsf and gradient adaptive convolution,'' in {\em Proceedings of the 31st ACM international conference on multimedia}, pp.~2446--2457, 2023.

\bibitem{li2018haze}
J.~Li, Q.~Hu, and M.~Ai, ``Haze and thin cloud removal via sphere model improved dark channel prior,'' {\em IEEE Geoscience and Remote Sensing Letters}, vol.~16, no.~3, pp.~472--476, 2019.

\bibitem{liu2017haze}
Q.~Liu, X.~Gao, L.~He, and W.~Lu, ``Haze removal for a single visible remote sensing image,'' {\em Signal Processing}, vol.~137, pp.~33--43, 2017.

\bibitem{ju2021ide}
M.~Ju, C.~Ding, W.~Ren, Y.~Yang, D.~Zhang, and Y.~J. Guo, ``Ide: Image dehazing and exposure using an enhanced atmospheric scattering model,'' {\em IEEE Transactions on Image Processing}, vol.~30, pp.~2180--2192, 2021.

\bibitem{he2023srd}
Y.~He, C.~Li, and T.~Bai, ``Remote sensing image haze removal based on superpixel,'' {\em Remote Sensing}, vol.~15, no.~19, p.~4680, 2023.

\bibitem{talebi2018nima}
H.~Talebi and P.~Milanfar, ``Nima: Neural image assessment,'' {\em IEEE Trans. Image Process.}, vol.~27, no.~8, pp.~3998--4011, 2018.

\bibitem{mittal2013making}
A.~Mittal, R.~Soundararajan, and A.~C. Bovik, ``Making a “completely blind” image quality analyzer,'' {\em IEEE Signal Processing Letters}, vol.~20, no.~3, pp.~209--212, 2013.

\bibitem{yang2015underwater}
M.~Yang and A.~Sowmya, ``An underwater color image quality evaluation metric,'' {\em IEEE Transactions on Image Processing}, vol.~24, no.~12, pp.~6062--6071, 2015.

\bibitem{he2026efficient}
L.~He, Z.~Yi, P.~Li, S.~Wang, C.~Chen, and M.~Lu, ``Efficient single image dehazing based on gradient line prior,'' {\em IEEE Transactions on Multimedia}, 2026.

\bibitem{huang2026haze}
C.~Huang, S.~Li, X.~Chen, J.~Liu, and D.~Li, ``Haze has many faces: Multi-domain haze style transfer for diverse haze removal,'' {\em Pattern Recognition}, p.~113094, 2026.

\bibitem{li2015nighttime}
Y.~Li, R.~T. Tan, and M.~S. Brown, ``Nighttime haze removal with glow and multiple light colors,'' in {\em Proceedings of the IEEE international conference on computer vision}, pp.~226--234, 2015.

\bibitem{varghese2024yolov8}
R.~Varghese and M.~Sambath, ``Yolov8: A novel object detection algorithm with enhanced performance and robustness,'' in {\em 2024 International conference on advances in data engineering and intelligent computing systems (ADICS)}, pp.~1--6, IEEE, 2024.

\bibitem{lowe2004distinctive}
D.~G. Lowe, ``Distinctive image features from scale-invariant keypoints,'' {\em International journal of computer vision}, vol.~60, no.~2, pp.~91--110, 2004.

\bibitem{lei2018superpixel}
T.~Lei, X.~Jia, Y.~Zhang, S.~Liu, H.~Meng, and A.~K. Nandi, ``Superpixel-based fast fuzzy c-means clustering for color image segmentation,'' {\em IEEE Transactions on Fuzzy Systems}, vol.~27, no.~9, pp.~1753--1766, 2018.

\end{thebibliography}

\begin{IEEEbiography}[{\includegraphics[width=1in,height=1.25in,clip,keepaspectratio]{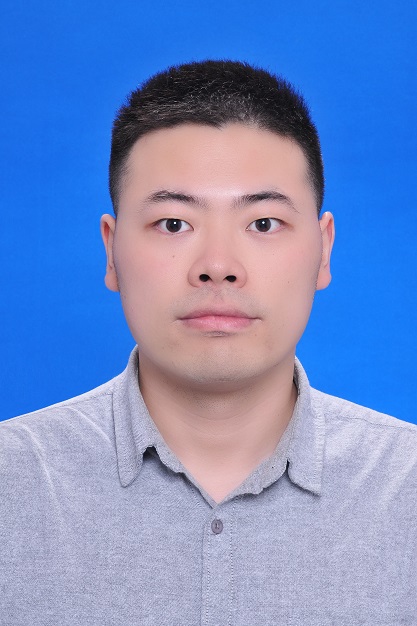}}]{Yun Liu}
received the M.S. and Ph.D. degrees in computer science and technology from Sichuan University, Chengdu, China,
in 2016 and 2019, respectively. He is currently an Associate
Professor with the College of Artificial Intelligence, Southwest
University, Chongqing, China. From 2024 to 2025, he was supported by China Scholarship Council and working with
Prof. Weisi Lin as a Postdoctoral Research Fellow with the College of Computing and Data Science, Nanyang Technological University, Singapore. He has published over 30 journal and conference papers as the first author or corresponding author, including IEEE Transactions on Image Processing, IEEE Transactions on Multimedia, IEEE Transactions on Circuits and Systems for Video Technology, IEEE Transactions on Intelligent Transportation Systems, ICCV, ACM Multimedia, and AAAI. His current research interests include image processing, computer vision, and deep learning.
\end{IEEEbiography}

\begin{IEEEbiography}
[{\includegraphics[width=1in,height=1.25in,clip,keepaspectratio]{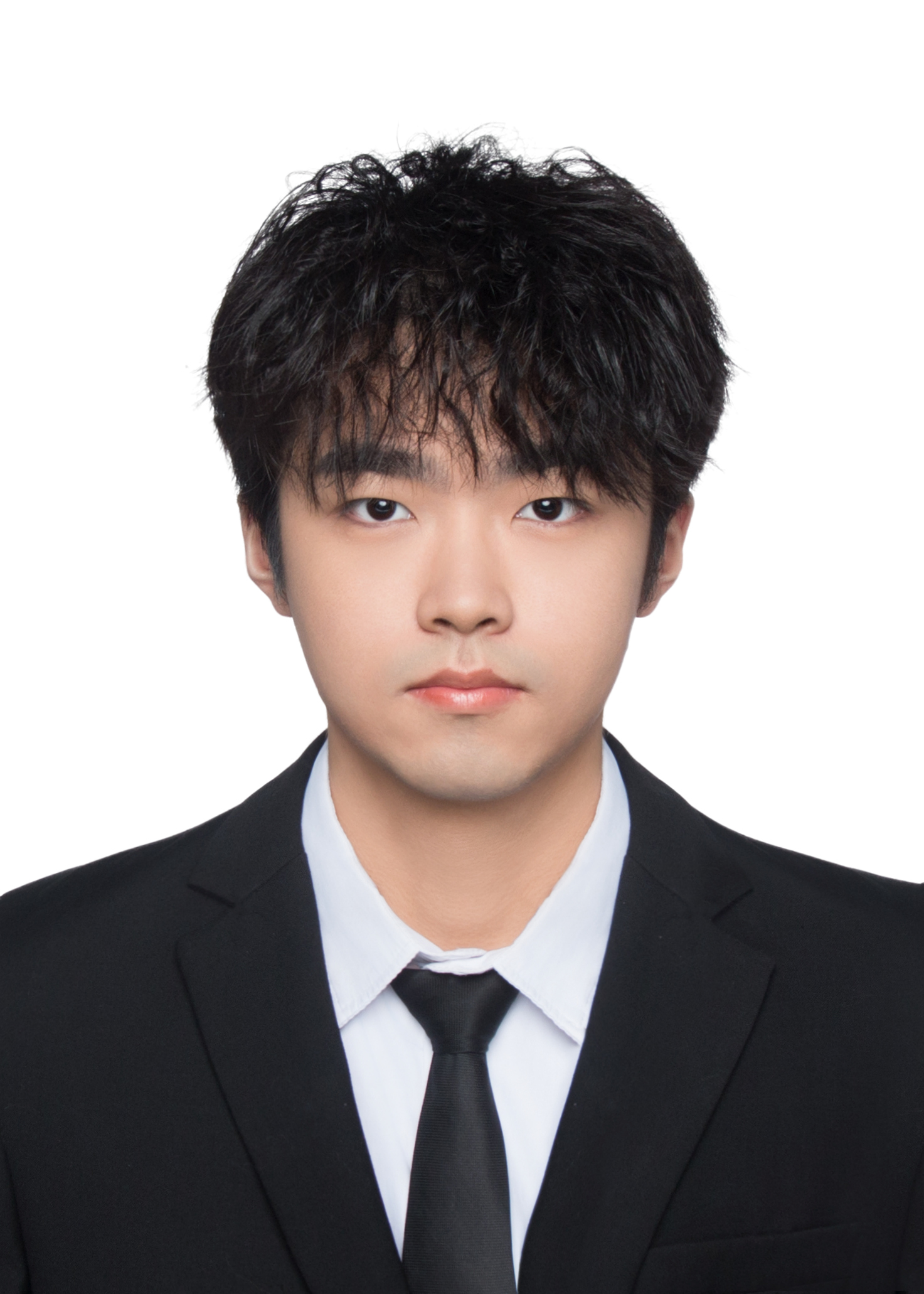}}]{Tao Li}
is currently pursuing the B.S.
degree in artificial intelligence with
the College of Artificial Intelligence, Southwest University. He has authored or coauthored research papers in IEEE Transactions on Image Processing, IEEE Transactions on Circuits and Systems for Video Technology, and AAAI. His current research interests include image
dehazing and deep learning.
\end{IEEEbiography}

\begin{IEEEbiography}
[{\includegraphics[width=1in,height=1.25in,clip,keepaspectratio]{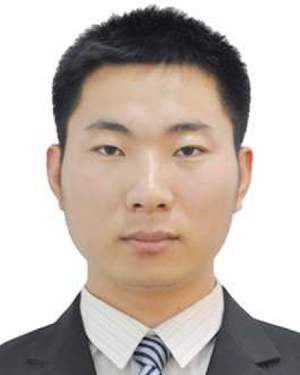}}]{Guanghui Yue} (Senior Member, IEEE) received the B.S. degree in communication engineering and the Ph.D. degree in information and communication engineering from Tianjin University, Tianjin, China, in 2014 and 2019, respectively. He was a Joint Ph.D. Student with the School of Computer Science and Engineering, Nanyang Technological University, Singapore, from September 2017 to January 2019. He is currently an Associate Professor with the School of Biomedical Engineering, Shenzhen University Medical School, Shenzhen University. His research interests include medical image analysis, bioelectrical signal processing, image quality assessment, 3D image visual discomfort prediction, pattern recognition, and machine learning.
\end{IEEEbiography}

\begin{IEEEbiography}[{\includegraphics[width=1in,height=1.25in,clip,keepaspectratio]{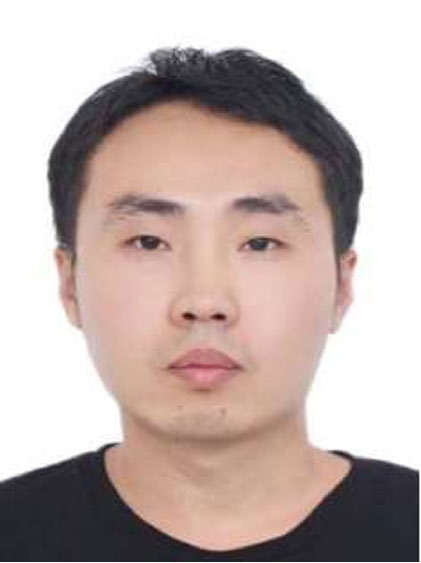}}]{Wenqi Ren} (Senior Member, IEEE)
received the Ph.D. degree from Tianjin University, Tianjin, China,
in 2017. From 2015 to 2016, he was supported
by China Scholarship Council and working with
Prof. Ming-Husan Yang as a Joint-Training Ph.D.
Student with the Electrical Engineering and Computer Science Department, University of California
at Merced. He is currently a Professor with the
School of Cyber Science and Technology, Sun Yatsen University, Shenzhen Campus, Shenzhen, China.
His research interests include image processing and
related high-level vision problems. He received the Tencent Rhino Bird Elite
Graduate Program Scholarship in 2017 and the MSRA Star Track Program
in 2018.
\end{IEEEbiography}

\begin{IEEEbiography}[{\includegraphics[width=1in,height=1.25in,clip,keepaspectratio]{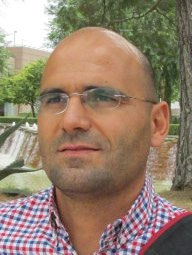}}]{Cosmin Ancuti}
received the Ph.D. degree from Hasselt University, Belgium,
in 2009. He was a Post-Doctoral Fellow with IMINDS and Intel Exascience Laboratory (IMEC), Leuven, Belgium, from 2010 to 2012, and a Research Fellow with the University Catholique of Louvain, Belgium, from 2015 to 2017. Hs is  currently a Full Professor at the Politehnica University of Timișoara (UPT), Romania. He has authored over 60 scientific papers, accumulating more than 9,000 citations and an h-index of 30. Professor Ancuti has served as Principal Investigator for five research projects and is a regular reviewer for leading computer vision conferences such as CVPR, ICCV, and ECCV. Since 2018, he has been a co-organizer of the IEEE CVPR NTIRE Workshop and the associated challenges on image dehazing. In recognition of his scientific contributions, he received the Gheorghe Cartianu Award of the Romanian Academy in 2020. His research interests include  image dehazing, underwater image enhancement, and low-light image enhancement.
\end{IEEEbiography}

\begin{IEEEbiography}[{\includegraphics[width=1in,height=1.25in,clip,keepaspectratio]{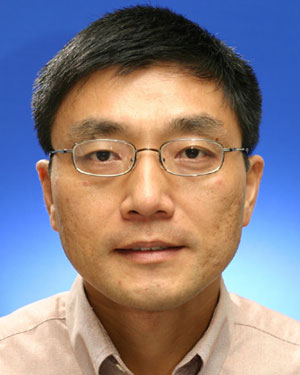}}]{Weisi Lin} (Fellow, IEEE) received the Ph.D.
degree from the King’s College, University of
London, U.K. He is currently a Professor with
the School of Computer Science and Engineering, Nanyang Technological University. His areas
of expertise include image processing, perceptual
signal modeling, video compression, and multimedia
communication, in which he has published over
200 journal articles, over 230 conference papers,
filed seven patents, and authored two books. He has
been an invited/panelist/keynote/tutorial speaker at
over 20 international conferences. He is a fellow of IET and an Honorary Fellow of Singapore Institute of Engineering Technologists. He has been the Technical Program Chair of IEEE ICME 2013, PCM 2012, QoMEX 2014, and IEEE VCIP 2017. He has been an Associate Editor of IEEE
TRANSACTIONS ON IMAGE PROCESSING, IEEE TRANSACTIONS ON CIRCUITS AND SYSTEMS FOR VIDEO TECHNOLOGY, IEEE TRANSACTIONS
ON MULTIMEDIA, and IEEE SIGNAL PROCESSING LETTERS. He was a Distinguished Lecturer of Asia-Pacific Signal and Information Processing
Association (APSIPA) from 2012 to 2013 and the IEEE Circuits and Systems
Society from 2016 to 2017.
\end{IEEEbiography}

\end{document}